\documentclass[11pt]{article}

\usepackage[final]{acl}

\usepackage{times}
\usepackage{latexsym}

\usepackage[T1]{fontenc}

\usepackage[utf8]{inputenc}

\usepackage{microtype}

\usepackage{booktabs}

\usepackage{inconsolata}

\usepackage{graphicx}

\usepackage{amsmath}
\usepackage{amssymb}
\usepackage{multirow}
\usepackage{xcolor}
\usepackage{url}
\usepackage{subcaption}

\makeatletter
\g@addto@macro\UrlBreaks{\do\A\do\B\do\C\do\D\do\E\do\F\do\G\do\H\do\I\do\J\do\K\do\L\do\M\do\N\do\O\do\P\do\Q\do\R\do\S\do\T\do\U\do\V\do\W\do\X\do\Y\do\Z\do\a\do\b\do\c\do\d\do\e\do\f\do\g\do\h\do\i\do\j\do\k\do\l\do\m\do\n\do\o\do\p\do\q\do\r\do\s\do\t\do\u\do\v\do\w\do\x\do\y\do\z}
\makeatother

\ifdefined\linenumbersep\fi

\graphicspath{{figures/}}

\title{Grounded Revision vs. Prior Injection: Probing Retrieval-Augmented Patent Claim Amendment}

\author{
  \textbf{Josepha Michiko Leo\textsuperscript{1}}\thanks{~Equal contribution.},
  \textbf{Hyun-seok Min\textsuperscript{2}}\footnotemark[1],
  \textbf{Yehoon Jang\textsuperscript{1}},
  \\
  \textbf{Irvan Zidny\textsuperscript{1}},
  \textbf{Jin-Woo Chung\textsuperscript{3}},
  \textbf{Sungchul Choi\textsuperscript{1,4}}\thanks{~Corresponding author.}
  \\
  \\
  \textsuperscript{1}Major in Industrial Data Science \& Engineering, \\
  Department of Industrial and Data Engineering, Pukyong National University \\
  \textsuperscript{2}Tomocube Inc. \quad \textsuperscript{3}Connectionary \quad \textsuperscript{4}Teamreboott Inc. \\
  {\small \texttt{\{jmichikoleo, jangyh0420, zidny4399\}@pukyong.ac.kr}} \\
  {\small \texttt{sc82.choi@pknu.ac.kr}, \texttt{min6284@gmail.com}, \texttt{jwchung@connectionary.io}}
}

\begin{document}
\maketitle
\begin{abstract}
Retrieval-augmented generation is widely used in professional writing, yet whether retrieval grounds revision or merely injects templates is rarely tested where ``correct'' has a definable meaning. Patent claim amendment supplies that signal: the examiner names the attacked limitation and cites prior art, providing per-case ground truth. We release three artifacts: (i) a corpus of 7,385 USPTO prosecution cases with XML-aligned pre/post claims, rejection, and cited prior art; (ii) a seven-probe battery comparing random and structural-match retrieval as two policies under a fixed prompt scaffold; (iii) a deterministic five-channel metric (C1--C3 and C5 in main, C4 supplementary) requiring no LLM evaluation. Across 9,600 pre-registered calls on four frontier LLMs (Claude Sonnet 4, Claude Haiku 4.5, GPT-5.4, GPT-4o-mini), no tested model exhibits \emph{detectable} classical prior-injection behavior; retrieval effects are small and direction-inconsistent between random and structural retrieval, and the null is unchanged under a dense (semantic) retriever, across retrieval depths $k\in\{1,3,5,10\}$, and under a paraphrase-sensitive grounding metric. Revision locality reveals a model-specific difference that the template channel misses. The four-cell taxonomy, which we treat as exploratory, leaves the prior-injector cell unoccupied.
\end{abstract}

\section{Introduction}
\begin{figure*}[!t]
\centering
\includegraphics[width=1\linewidth]{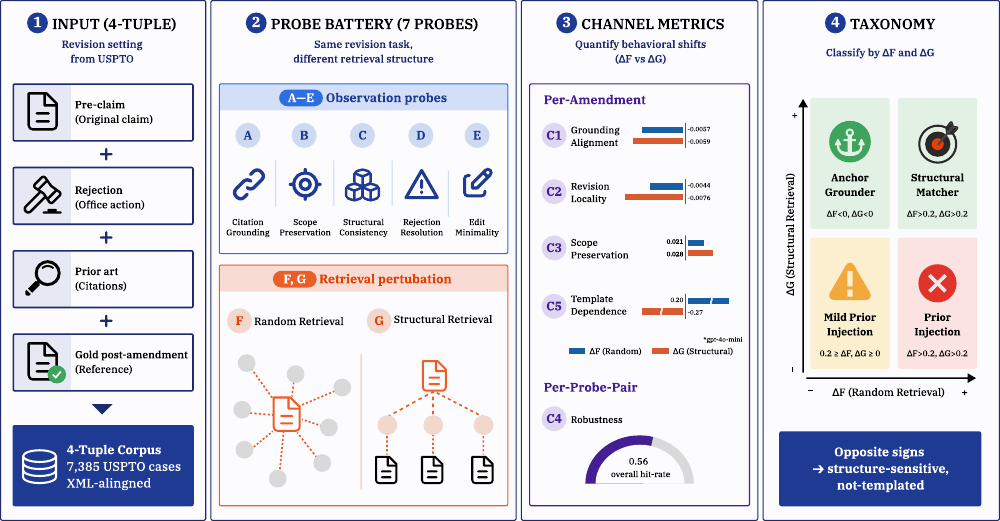}
\caption{Does the model address the examiner-attacked limitation (top, grounded revision) or recycle generic patent phrasing (bottom, prior injection)? Our probe battery (\S\ref{sec:probes}) and metric (\S\ref{sec:metric}) distinguish the two without an LLM judge.}
\label{fig:Figure1}
\end{figure*}

Automatic generation of structured technical documents is expanding faster than its evaluation infrastructure. Recent end-to-end systems generate hypotheses \citep{gottweis2025}, write manuscripts \citep{song2026}, and conduct their own peer review \citep{lu2026}: \citet{lu2026} report a generated manuscript passing workshop peer review, and more than 120 computer-generated papers were retracted from IEEE and Springer journals over a decade ago \citep{vannoorden2014}. By comparison, code generation matured only after HumanEval \citep{chen2021} and SWE-bench \citep{jimenez2024} introduced structurally rigorous evaluation tied to executable ground truth.

Patents provide that anchor: examiners apply shared legal standards (35 U.S.C. \S102 novelty, \S103 non-obviousness, \S112 enablement) under the standardized Manual of Patent Examining Procedure (MPEP), creating an experimentally controlled setting in which correctness is legally definable. The USPTO receives over 612,000 new utility, plant, and reissue patent applications annually and processes more than 137,000 Requests for Continued Examination \citep{uspto2025}. Patent prosecution is a paradigm case of grounded revision: each amendment is paired with an examiner-identified attacked limitation and the on-record passage of a cited reference, an alignment most revision corpora lack.

Commercial systems such as PatentGPT (Patsnap) and IPRally Drafter, along with a growing set of startups, market retrieval-augmented amendment generation on the premise that retrieving prior examples grounds the output; to our knowledge these are vendor product claims advanced without peer-reviewed evaluation, which is part of what motivates an independent test. Whether retrieval substantively grounds outputs in structural similarity, or serves more as an undifferentiated prior-injection anchor whose effect is independent of the specific items retrieved, is an open empirical question.

Academic work has progressed in parallel but has not tested whether retrieval substantively engages with the rejection or merely produces surface-plausible prose. Patent-CR \citep{jiang2024} releases 22,606 draft-to-grant pairs but evaluates only with surface-level similarity. PANORAMA \citep{lim2025} enumerates claim revision without evaluation. PEDANTIC \citep{knappich2025} and PILOT-Bench \citep{jang2025} classify \S112 indefiniteness and rejection-issue types respectively, without generation. None tests retrieval's role in grounded revision under a controlled policy contrast. Plausibility is cheap: any competent generator produces patent-sounding text, but grounded revision additionally requires that the edit target the attacked limitation, differentiate from the cited mechanism, and neither over-narrow nor recycle boilerplate. These failure modes leave textual fingerprints (MPEP \S714), and we exploit them to build a probe-based evaluation protocol.

\paragraph{Contributions.} The paper contributes three coupled components.

\textbf{Corpus (\S\ref{corpus}).} We release 7,385 four-tuples (pre-amendment claim, rejection, cited prior art, post-amendment claim) aligned at the XML level from the USPTO Open Data Portal, supporting a stratified 100-case test cohort (\S\ref{sec:cohort}) and sampling-bias robustness checks (Appendix~\ref{app:replication}).

\textbf{Probe battery (\S\ref{sec:probes}).} We design seven probes: five observational (A--E) and two retrieval-intervention (F, G) that realize a random-versus-structural policy contrast under a fixed prompt scaffold (Figure~\ref{fig:Figure1}). The observational axis makes the F-versus-G comparison interpretable rather than merely measurable.

\textbf{Metric (\S\ref{sec:metric}).} We define a five-channel deterministic metric with no LLM in the evaluation loop. C1--C3 and C5 are per-amendment channels and carry the main-paper verdicts; C4 (robustness) is defined on probe pairs and is reported in supplementary tables (\S\ref{sec:channels}).

Together these test three pre-registered hypotheses (H1--H3, \S\ref{sec:hypotheses}) in a 2$\times$2 factorial across Claude and OpenAI families at flagship and smaller tiers.

\textbf{Scope}. We study applicant-side amendment generation, i.e., given a rejection, generate the post-amendment claim.

\textbf{Preview of findings.}
Across the four tested LLMs, the prior-injector cell of our retrieval-mechanism taxonomy turned out to be empty: random and structural retrieval produce small, often opposite-sign shifts in template dependence rather than the uniform inflation that the most pessimistic grounding-as-template-recycling account would predict. The framework's value does not rest on that null: it registers model-specific differences on revision locality (most clearly for GPT-5.4) that a single-channel test would miss, and, treating the four-cell taxonomy as exploratory, locates each tested model within it.

\section{Related Work}
\subsection{Patent NLP}
The closest prior work is Patent-CR \citep{jiang2024}, which released 22,606 rejected-to-granted claim pairs evaluated with BLEU, ROUGE, BERTScore, G-Eval, and a five-axis expert rating. The evaluation scores pair fidelity but does not expose the examiner's rejection or the cited prior art, nor does it contrast retrieval policies under a controlled scaffold. We add aligned rejection and prior-art channels to the input, and switch from fidelity scoring to a probe-based policy-contrast evaluation. Other patent NLP work sits outside a controlled retrieval contrast for structural reasons. BIGPATENT \citep{sharma2019} frames patent text as an abstractive summarization corpus; PATENTWRITER \citep{shomee2025} and AutoPatent \citep{wang2024} generate claims or full specifications from non-conditional inputs; and generation quality is assessed by evaluation-focused work, including \citet{lee2023}, \citet{jiang2025}, and PatentScore \citep{yoo2025}. On the classification side, PEDANTIC \citep{knappich2025} labels \S112 definiteness, PANORAMA \citep{lim2025} enumerates claim revisions without evaluation, PILOT-Bench \citep{jang2025} classifies rejection-issue types, and \citet{shi2024} predicts examination outcomes. In none of these does the retrieval channel enter the input aligned with the rejection that retrieved exemplars would need to address, so none admits the policy-contrast test we run. Related resources also align examiner rejections with claims: ClaimBrush \citep{claimbrush2024} pairs pre/post claims with Office-Action metadata for Japanese filings, and Tree-of-Claims \citep{treeofclaims2025} and the USPTO Office Action Research Dataset \citep{oard2017} align cited prior art with rejected claims. Our contribution is more delimited: we align the pre- and post-amendment claim text, the rejection, and the cited prior art as four-tuples at the XML level, with an application-number rebuild index, and frame them as model input, which is what allows Probes F and G to be interpreted as a contrast between two retrieval policies inserted into the same prompt scaffold. InstructPatentGPT \citep{instructpatentgpt2024} similarly uses Office Actions as a training signal rather than as a controlled retrieval contrast.

\subsection{RAG evaluation and behavioral probing}
RAG evaluation has progressed from retrieval-augmented pre-training \citep{guu2020, lewis2020} through few-shot retrieval augmentation \citep{izacard2022} and self-reflective retrieval \citep{asai2023}, with automated evaluation metrics \citep{es2024} and heterogeneous zero-shot retrieval benchmarks \citep{thakur2021}, but remains predominantly correlational. We borrow intervention logic from behavioral probing \citep{elazar2021, ribeiro2020, vig2020}: our observational probes (input perturbations, detailed as A--E in \S\ref{sec:probes}) follow CheckList, and our two retrieval interventions (random versus structurally-matched retrieval, F and G in \S\ref{sec:probes}) are policy-contrast interventions on the retrieval channel, in the design tradition of \citet{vig2020} but at the level of a comparison between two retrieval policies inserted into a fixed prompt scaffold rather than full causal identification of retrieval per se. Work on retrieval noise shows that even random or irrelevant passages can shift generation substantially \citep{cuconasu2024, fang2024raat}, and that context position matters \citep{liu2024lost}; correlational RAG benchmarks, however, cannot separate similarity-driven grounding from generic context injection, which our F-versus-G contrast isolates. The patent-amendment setting addresses this confound at the corpus level: every test case carries an examiner-identified attacked limitation and an on-record cited reference, so the input is fully specified before retrieval is added, and any shift produced by F or G is attributable to the retrieval-policy intervention. This enables the framework to place models in the four-cell taxonomy of \S\ref{sec:taxonomy} rather than only ranking them on a single retrieval-quality scalar.

\section{Patent Amendment 4-Tuple Corpus}
\label{corpus}

We release a corpus of 7,385 prosecution cases in which every amendment is aligned at the XML level with its triggering rejection, the cited prior art, and the pre- and post-amendment claim text. The scale supports the 100-case test cohort (\S\ref{sec:cohort}), a 4,221-case retrieval pool for Probes F and G, and a sampling-bias replication reserve (Appendix~\ref{app:replication}).
\subsection{Source and alignment}
\label{sec:corpus-source}
We start from the PILOT-Bench PTAB-appeal subset \citep{jang2025} of 13,749 cases and enrich each case via the USPTO Open Data Portal (ODP). For each case we resolve its application number, retrieve the file-wrapper document list, and identify the first non-final rejection together with the immediately preceding Claims filing (pre-amendment state) and the first Claims filing after the rejection (post-amendment state). For each pre/post claim pair we compute a per-claim difference status (kept, modified, new, cancelled) using claim-number alignment, and for modified claims a SequenceMatcher-based similarity ratio plus the list of added and removed spans. The difference is the ground truth against which the generated amendments are scored (\S\ref{sec:metric}).

\subsection{Summary statistics}
\label{sec:corpus-stats}

Two corpus-level findings support the paper's framing. First, amendment is predominantly surgical: 64.5\% of per-claim actions are modifications at median similarity 0.933 ($\sim$7\% character-level edit), with cancellations and new claims accounting for another 32.5\%. Second, the rejection-to-amendment linkage holds at the parse level: 94.5\% of scorable cases show overlap between rejected and post-amendment-modified claim numbers. Of the 7,385 parseable triples, 5,755 admit a case-level C2 denominator (\S\ref{sec:metric}). The first-rejection statute breakdown is dominated by \S102 (2{,}337 cases), with \S112 (1{,}245), \S101 (950), and \S103 (738) accounting for the remainder. Full corpus statistics are reported in Appendix~\ref{app:funnel}.

\subsection{Release}
\label{sec:corpus-release}
We release the parsed JSONL corpus and parsing code under a permissive open license; an application-number index permits zero-cost reconstruction from USPTO ODP, whose public data have no copyright restriction. Our corpus is the only one of Patent-CR, PANORAMA, PEDANTIC, and PILOT-Bench that aligns all four elements (rejection context, cited prior art, pre/post claim pair, and amendment diff) at the XML level. Patent-CR provides the pre/post pair but omits the rejection and prior-art channels; PANORAMA, PEDANTIC, and PILOT-Bench target judgment, classification, or upstream retrieval rather than amendment generation.

\section{Method}
\label{sec:method}

Our method has two components. The probe battery (\S\ref{sec:probes}) defines seven controlled input manipulations: five observational baselines (A--E) plus a two-policy retrieval contrast (F vs G) inserted into a fixed prompt scaffold, with the F/G comparison interpreted against the observational baseline rather than as unconditional causal identification of retrieval. The five-channel deterministic metric (\S\ref{sec:metric}) scores every probe condition on common axes without using an LLM as judge.

\subsection{Probe Battery}
\label{sec:probes}
We design two probe families read jointly. Observational probes A--E characterize each LLM's baseline amendment signature under input perturbations. Retrieval probes F and G intervene on the retrieval channel via a controlled random-versus-structural policy contrast under a fixed insertion template. The observational axis supplies the interpretive frame against which the retrieval-policy effect is measured, so an F-versus-G shift is read relative to a characterized baseline rather than an undifferentiated mean. All probes share a common input template (pre-amendment claim, rejection rationale, cited prior art) and differ only in what is manipulated; each probe's expected effect on the five-channel score (\S\ref{sec:metric}) is fixed in advance.
\paragraph{Observational probes (A--E).}
\label{sec:obs-probes}
\textbf{Probe A (Claim Truncation)} modifies a single decisive limitation in the rejected independent claim by deletion, addition, paraphrase, or antonym swap. A grounded model follows the perturbation in C1; a model that ignores the rejection signal will show C1 invariance.

\textbf{Probe B (Rationale Shuffling)} compares the original examiner findings against a version with one sentence replaced by a semantically inconsistent fragment or with sentence order permuted. A rationale-grounded model holds C1 stable across the coherence break, while a surface tracker degrades.

\textbf{Probe C (Decoy Citation)} substitutes either a wording-similar reference that does not teach the mechanism, or a wording-dissimilar reference that does, separating lexical from mechanistic engagement with the cited art.

\textbf{Probe D (Boilerplate Injection)} prepends canonical patent boilerplate to the drafting context. A grounded model treats the prepended phrasing as scaffolding and leaves C5 and C3 largely unchanged; a prior-injection-prone model inflates C5.

\textbf{Probe E (Drafting Hint)} adds one of four task hints: ``make minimal amendment'', ``preserve scope'', ``focus only on novelty'', or ``avoid unnecessary narrowing''. A model with stable instruction-following should produce measurable shifts on C2 and C3 in the directions the hints predict.

\paragraph{Retrieval probes (F, G).}
\label{sec:ret-probes}
\textbf{Probe F (Random Retrieval)} retrieves $k$ past amendments from the corpus pool (excluding the test case and its ancestors), selected uniformly at random, and prepends them as context. Random retrieval realizes the prior-injection condition in its strongest form: any effect cannot be attributed to similarity-driven grounding because selection is similarity-agnostic by construction.

\textbf{Probe G (Structural-Match Retrieval)} retrieves $k$ past amendments matched on (i) statute section, (ii) statute subsection, and (iii) a coarse limitation-pattern feature derived from the attacked claim. The deterministic match isolates the structural-similarity signal: any shift attributable to the match itself must surface in the F-versus-G delta.

Both probes use the same $k$ (default 3; retrieval-depth sensitivity in Appendix~\ref{app:ksweep}) and the same prompt-insertion template; G uses a deterministic feature-based matcher rather than a dense retriever to keep the F and G contrast interpretable as a structural-similarity test (Appendix~\ref{app:retrieval}). Because F, G, and the no-retrieval baseline all sit inside the same prompt scaffold, F-versus-G identifies the effect of switching retrieval policy under a fixed scaffold rather than the unconditional effect of adding retrieval. The empty prior-injector cell that the framework finds in \S\ref{sec:taxonomy} is therefore a finding about the deterministic structural-similarity policy realized by Probe G, not about retrieval-augmented generation in general; dense-retriever configurations would constitute a different point in the same policy-contrast design and can be evaluated with the released framework without modification.

\subsection{Five-Channel Evaluation Metric}
\label{sec:metric}
Using an LLM as evaluator embeds the evaluator's biases into the score, a concern made empirical by our findings (\S\ref{sec:results}): the four frontier models cluster into distinct behavioral patterns under retrieval perturbation, so selecting any one as judge would install that model's signature as the evaluation axis. Therefore, we adopt a deterministic five-channel metric. Each channel is a function of text, parser output, and the XML gold post-amendment claim; channels are designed to be jointly informative, and we verify their near-independence empirically (\S\ref{sec:methodology-validity}).
\paragraph{Channels.}
\label{sec:channels}
\textbf{C1 (grounding alignment)} measures the overlap between limitations modified in the generated amendment and limitations named as rejected in the office action, computed at limitation-token level after claim-number alignment. A high C1 means the model is editing the same limitation the examiner attacked.

\textbf{C2 (revision locality)} is the ratio $\mathrm{editdist}(\mathrm{gen}, \mathrm{pre}) / \mathrm{editdist}(\mathrm{gold\_post}, \mathrm{pre})$; a value of 1 indicates the model edited at the same scale as the gold amendment, $\ll 1$ indicates under-editing, and $\gg 1$ indicates over-rewriting. C2 catches timidity and over-rewriting as numerically distinct rather than collapsing them into a single similarity score.

\textbf{C3 (scope preservation)} is the Jaccard overlap of noun phrases between the generated amendment and the original claim's invention core, after limitation-removal normalization; reduction indicates over-narrowing. The normalization isolates true scope drift from the mechanical narrowing that any limitation deletion induces.

\textbf{C4 (robustness)} measures, for any pair of probe conditions, whether the amendment shifts in the predicted direction, scored per probe and averaged within a model. C4 is defined on probe pairs rather than on individual amendments, so we report it in supplementary material (per-pair tables released alongside Appendix Table~\ref{tab:probe-deltas}) to keep the main-paper tables dimensionally consistent with the per-amendment channels C1--C3 and C5. The headline verdicts in \S\ref{sec:verdicts} and the taxonomy in \S\ref{sec:taxonomy} are therefore computed on the per-amendment channels; C4 enters only as a supplementary directional-hit-rate check. On the 100-case cohort, the grand C4 across the seven pre-registered (probe, channel) pairs is $0.56$ ($n{=}2{,}276$ case-model contrasts; per-pair table in Appendix~\ref{app:c4}).
 
\textbf{C5 (template dependence)} is the frequency of canonical amendment phrases mined from the retrieval pool, normalized per 1,000 characters of generated claim text. C5 is the central channel for adjudicating the prior-injection hypothesis: if retrieval inflates the rate of canonical phrases, C5 rises in lockstep. We emphasize that C5's absolute level is not a measure of prior injection: canonical phrasing is normal in competent drafting, so a high C5 may reflect legitimate convention. Prior injection is diagnosed only from $\Delta$C5, the retrieval-induced change over the no-retrieval baseline for the same case and model, so convention present at baseline is differenced out and only retrieval-added phrasing is attributed to injection. C5 remains a proxy that cannot alone separate appropriate phrasing from recycling, and a lower C5 need not indicate better grounding; criterion validation is deferred to the post-review study (\S\ref{sec:metric-validity}).

\paragraph{Validity.}
\label{sec:metric-validity}
The metric design follows the multitrait-multimethod logic of \citet{campbell1959}: each channel ties to a specific XML-record feature rather than an LLM judgment. Three legs support the metric: (i) within-case construct checks where each channel moves in the direction a patent practitioner would predict (Appendix~\ref{app:construct-checks}); (ii) empirical near-independence pooled across 3{,}062 (case, condition, model) triples, all off-diagonal correlations $|\rho|\leq 0.29$, below the pre-registered $|\rho|>0.7$ flag threshold (Table~\ref{tab:correlations}); (iii) a 30--50-case double-rated attorney study committed post-review (a 10-case pre-release sanity sample showed all four channels moving in the direction the reading attorney predicted; Appendix~\ref{app:attorney}). The construct-check and channel-independence legs carry the metric-validity argument in this paper; the formal $\kappa$ leg is deferred to the post-review study.

\section{Experiments}
\label{sec:experiments}
We fix the entire experimental protocol before any model call is made, following the pre-registration discipline recommended for probe-based evaluations \citep{vanmiltenburg2021}.
\subsection{Cohort and Retrieval Pool}
\label{sec:cohort}
The 100-case test cohort is constructed via joint stratified sampling across six axes (subdecision outcome, rejection statute, technology center, amendment pattern, XML format era, decision year bin) matching the marginal distributions of the 4,270-case eligibility pool while guaranteeing minimum cell sizes (Appendix~\ref{app:cohort}). The remaining 7,285 cases, after filtering for resolvable attacked-claim alignment, yield a 4,221-case retrieval pool indexed by (statute section, statute subsection, limitation pattern), used by Probes F and G. The cohort and pool are non-overlapping by construction.

\subsection{Model matrix}
\label{sec:model-matrix}
We run a 2$\times$2 factorial across vendor family and model tier. Stage 1 (Claude Sonnet 4 and GPT-5.4 flagships) runs the full probe battery. Stage 2 extends to Claude Haiku 4.5 and GPT-4o-mini, conditional on Stage 1 effect-size criteria detailed in Appendix~\ref{app:stage2}. Each case is scored under 8 conditions $\times$ 3 replicates (24 evaluations per case per model), giving within-model paired comparisons with $>80\%$ power at $d=0.3$, $\alpha=0.05$. Inference settings: temperature 0.3 for all models; three independent replicates per (case, probe, model) with separate sampling seeds; retrieval $k=3$ (retrieval-depth sensitivity over $k\in\{1,3,5,10\}$ in Appendix~\ref{app:ksweep}).

\subsection{Pre-registered hypotheses}
\label{sec:hypotheses}

\textbf{H1 (retrieval as prior-injection anchor)}: Probe F (Random Retrieval) increases template dependence (C5) relative to baseline by at least 0.2 and does not improve grounding alignment (C1) by more than 0.1. This is the strongest commercial concern about retrieval-augmented drafting in operational form: if random exemplars inflate template phrasing without sharpening grounding, retrieval is acting as a vendor-agnostic boilerplate anchor.

\noindent\textbf{H2 (similarity as second-order)}: Probes F (random) and G (structural) yield statistically indistinguishable shifts in C1, C2, and C5 ($|\Delta_F - \Delta_G| \leq 0.1$ on each channel, or same sign with no significant ranking). H2 isolates the marginal value of structural matching: if F and G are observationally equivalent, the structural-similarity layer adds nothing the random baseline does not already provide.

\noindent\textbf{H3 (template-recycling inflation)}: retrieval does not reduce template dependence below baseline on either probe; equivalently, the grounded-revision hypothesis that retrieval reduces generic phrasing is not supported. H3 closes the loop in the opposite direction from H1, ruling out the optimistic ``retrieval makes models write more like experts'' story when both probes leave C5 at or above baseline.

\noindent The full $3 \times 3$ outcome space (each hypothesis $\in$ \{supported, rejected\}) is addressed in \S\ref{sec:taxonomy}.

\section{Results and Analysis}
\label{sec:results}
We ran the benchmark on four frontier LLMs (Claude Sonnet 4, Claude Haiku 4.5, GPT-5.4, GPT-4o-mini) across 100 cohort cases $\times$ 8 conditions $\times$ 3 replicates = 9,600 model calls. Parse-success rate was 97.4\% overall (93.0\% on GPT-4o-mini to 100\% on GPT-5.4). Per-response scores are aggregated as per-case mean across replicates, then median across cases within (model, condition).

\subsection{Pre-registered hypothesis verdicts}
\label{sec:verdicts}

\begin{table}[t]
    \centering
    \caption{Pre-registered hypothesis verdicts on C5 (canonical-phrase rate per 1{,}000 characters, $N{=}300$ per cell = 100 cases $\times$ 3 replicates). Y: supported; N: not supported; $\sim$: at-boundary, see \S\ref{sec:verdicts} for the boundary-handling rule.}
    \label{tab:verdicts}
    \resizebox{\columnwidth}{!}{%
    \begin{tabular}{@{}lcccccc@{}}
    \toprule
    Model & base & $\Delta$F & $\Delta$G & H1 & H2 & H3 \\
    \midrule
    Claude Sonnet 4   & 3.99 & $-0.16$ & $-0.13$ & N      & \textbf{Y} & N          \\
    Claude Haiku 4.5  & 3.40 & $+0.19$ & $+0.03$ & N      & N          & \textbf{Y} \\
    GPT-5.4           & 3.73 & $-0.12$ & $+0.06$ & N      & N          & N          \\
    GPT-4o-mini       & 3.14 & $+0.20$ & $-0.27$ & $\sim$ & N          & N          \\
    \bottomrule
    \end{tabular}%
    }
\end{table}

\textit{No model shows detectable prior injection.} H1 (retrieval-as-prior-injection) is not supported across all four models: $\Delta$C5 falls below the pre-registered 0.2 threshold in every case (max $+0.20$ on GPT-4o-mini, at the threshold). Under the pre-registered convention the threshold is the supported-side boundary, so $\Delta\mathrm{C5}=+0.20$ is marked ``$\sim$'' rather than ``Y''. Paired Wilcoxon signed-rank tests with Holm--Bonferroni correction and bootstrap 95\% CIs on the headline $\Delta$C5 values (Appendix~\ref{app:significance}) find no significant shift: 0 of 8 baseline-versus-retrieval comparisons (F and G across four models) reach uncorrected significance, and none survives correction. A two-one-sided-tests (TOST) equivalence check against the pre-registered $\pm 0.2$ margin further establishes that the median $\Delta$C5 is statistically equivalent to zero for all four models under both probes, so the GPT-4o-mini boundary is a tested equivalence rather than only an adjacency to threshold. We emphasize that equivalence within the pre-registered $\pm 0.2$ margin bounds the effect but does not establish its absence, and in particular does not exclude a true shift lying just below $+0.2$; we therefore report no \emph{detectable} prior injection at this sample size and margin rather than its absence.
Because any fixed cutoff is arbitrary, the verdict does not rest on a point estimate crossing $0.2$: no $\Delta$C5 is significant or distinguishable from zero (Appendix~\ref{app:significance}), so no model shows a \emph{significant} increase at any threshold in $0.15$--$0.25$, though the Haiku~4.5 and GPT-4o-mini point estimates sit near $0.2$.
H2 is uniquely supported by Sonnet 4: both F and G produce same-sign shifts within 0.1 (both modestly negative). The other three models show opposite-sign effects (GPT-5.4, GPT-4o-mini) or disparate magnitudes (Haiku 4.5); the pre-registered 0.10 bound is a strict bound, but H2 is jointly defined with the no-significant-ranking clause, so Haiku's $0.16$ gap fails the joint criterion ordinally as well and is recorded as ``N'' rather than at-boundary ``$\sim$''. H3 holds only for Haiku 4.5; Sonnet 4 and GPT-4o-mini each reduce template dependence under at least one retrieval condition. Per-case effects on C5 reach $\pm 2$ to $8$ even where aggregate medians stay sub-threshold; two worked examples are in Appendix~\ref{app:case-studies}. GPT-4o-mini over-edits at baseline (C2=1.46), and smaller-tier models show stronger Probe~A grounding shifts than flagships (full probe$\times$channel deltas in Appendix Table~\ref{tab:probe-deltas}).

\subsection{Model divergence on revision locality}
\label{sec:family-divergence}
Although H1 is null on the template channel, the benchmark registers model-specific differences on revision locality (C2). Table~\ref{tab:c2-family} shows retrieval reducing edit magnitude in both Claude models and in GPT-4o-mini, while GPT-5.4 slightly increases it, so the pattern is GPT-5.4 as the lone outlier rather than a clean Claude-versus-GPT split. We therefore treat this as a model-specific, exploratory observation rather than a vendor-level finding. Under paired Wilcoxon tests, 3 of 8 C2 comparisons are nominally significant (Haiku~4.5 F and G, Sonnet~4 F), all decreases, and none survives Holm correction (Appendix~\ref{app:significance}), so the effect is carried by Haiku~4.5. The point that a shift invisible on C5 surfaces on C2 still holds, showing the battery is not simply insensitive; we do not, however, claim a vendor-aligned family split.
\begin{table}[!htbp]
\centering
\small
\setlength{\tabcolsep}{5pt}
\caption{Revision-locality (C2) effects under random (F) and
structural (G) retrieval, per model.}
\label{tab:c2-family}
\begin{tabular*}{\columnwidth}{@{\extracolsep{\fill}}lccc@{}}
\toprule
Model            & base C2 & $\Delta$F & $\Delta$G \\
\midrule
Claude Sonnet~4  & 0.617   & $-0.074$  & $-0.013$  \\
Claude Haiku~4.5 & 1.147   & $-0.174$  & $-0.189$  \\
GPT-5.4          & 0.789   & $+0.016$  & $+0.058$  \\
GPT-4o-mini      & 1.459   & $-0.044$  & $-0.076$  \\
\bottomrule
\end{tabular*}
\end{table}

\subsection{Where the evidence places retrieval}
\label{sec:taxonomy}

\begin{figure}[t]
\centering
\includegraphics[width=\linewidth]{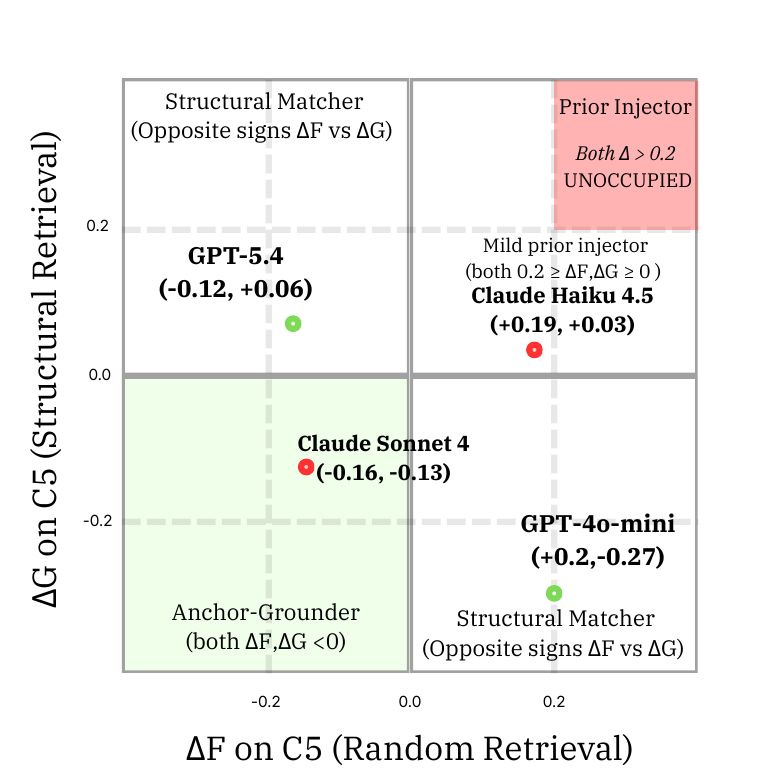}
\caption{Four retrieval-mechanism cells in $(\Delta_F, \Delta_G)$ space on C5. Coordinates from Table~\ref{tab:verdicts}: Sonnet 4 $(-0.16, -0.13)$ anchor-grounder; GPT-5.4 $(-0.12, +0.06)$ and GPT-4o-mini $(+0.20, -0.27)$ structural-matcher (opposite-sign F vs G); Haiku 4.5 $(+0.19, +0.03)$ inert / mildly prior-injecting. The prior-injector cell (both $\Delta > +0.2$) is unoccupied across all four models tested in this four-model, one-corpus, $k{=}3$, deterministic-structural-matcher configuration.}
\label{fig:taxonomy}
\vspace{-10pt}
\end{figure}

The pre-registered $2\times2\times3$ framework places four retrieval-mechanism accounts in play. The main experiment resolves them as follows, with each cell tagged by its $(\Delta_F, \Delta_G)$ coordinate on C5 from Table~\ref{tab:verdicts}.
\textbf{(a) Anchor-grounder} (H1 rejected, H3 rejected, retrieval reduces templateness without inflating alignment): Claude Sonnet 4 at $(-0.16, -0.13)$.
\textbf{(b) Prior injector} (both $\Delta_F$ and $\Delta_G$ above $+0.2$): not observed at pre-registered effect size.
\textbf{(c) Structural matcher} ($F \neq G$, H2 rejected): GPT-5.4 at $(-0.12, +0.06)$ (sign flip) and GPT-4o-mini at $(+0.20, -0.27)$ (0.47 directional gap).
\textbf{(d) Inert / mild prior injector} (H1 below threshold, H2 mild): Claude Haiku 4.5 at $(+0.19, +0.03)$ (Figure~\ref{fig:taxonomy}).
The ``$\sim$'' verdict for GPT-4o-mini on H1 and its cell-(c) assignment are complementary projections of the same point: cell-(b) membership requires the joint $(\Delta_F, \Delta_G)$ pair to clear threshold with consistent sign. This joint criterion is a post-hoc interpretive overlay, specified at the taxonomy step once the headline cell turned out empty, rather than a separately pre-registered threshold.
Haiku 4.5's cell-(d) tag follows the same overlay: H1 fails strictly and H2 fails with $|\Delta_F - \Delta_G|=0.16$ (above 0.1, same sign), so the F-versus-G pair is mild-but-not-equivalent rather than the same-sign-within-0.1 signature of cell (a).

\paragraph{Anchoring F and G against the Probe D boilerplate baseline.}
The cell assignments acquire interpretive weight when read against the observational deltas of Probes A--E for the same model (Appendix Table~\ref{tab:probe-deltas}). Probe D (Boilerplate Injection) provides the closest in-corpus reference for a generic prepended-context effect: on Sonnet 4 it moves $\Delta\mathrm{C5}$ by $+0.03$, on Haiku 4.5 by $-0.09$, on GPT-5.4 by $+0.07$, and on GPT-4o-mini by $-0.07$. 
For three of four models (Sonnet 4, GPT-5.4, GPT-4o-mini) the post-hoc descriptive gap $|\Delta_F|-|\Delta_D|$ is bounded at $0.13$; we use absolute values here because $\Delta_F$ and $\Delta_D$ disagree in sign on these models, so the comparison is one of prepended-context dosage rather than signed direction. At this dosage, random retrieval anchors C1/C2 toward the boilerplate prior rather than behaving qualitatively differently. Haiku 4.5 is the per-model exception, with $\Delta_F = +0.19$ exceeding $|\Delta_D| = 0.09$ (cell-(d) assignment; full per-model breakdown in Appendix~\ref{app:probe-d-anchor}). Probe C (Decoy Citation) similarly bounds the scale of mechanism-similarity confusion that any retrieval channel could induce, providing the C1 reference against which the near-zero $\Delta\mathrm{C1}$ values under F and G are read as null rather than as insensitivity.

\paragraph{\textbf{Distinguishing anchor-grounder from inert.}}Sonnet 4 (cell a) and Haiku 4.5 (cell d) are separated by two signals: the F-versus-G C5 sign pattern and C2. Sonnet 4 has same-sign $\Delta_F$, $\Delta_G$ within 0.1 (H2-supported), with small same-signed C2 shifts. Haiku 4.5 has $\Delta_F=+0.19$, $\Delta_G=+0.03$ (H2 rejected), with the largest absolute $\Delta\mathrm{C2}$ in the matrix ($-0.174$, $-0.189$).

\paragraph{Small, direction-inconsistent effects.} The modal retrieval behavior is therefore (c) or (d): small in magnitude and often direction-inconsistent between F and G. Discrimination on revision locality (C2; \S\ref{sec:family-divergence}) confirms that the framework detects retrieval-policy effects where they occur rather than being simply insensitive. Sonnet 4's H2 uniqueness ($F \approx G$, both slightly reducing template phrases) is the behavioral pattern most aligned with the originally conjectured grounded-revision mode.

\subsection{Benchmark methodology and metric validity}
\label{sec:methodology-validity}

\textit{Small samples mislead.} A smoke run at $N{=}10$ per cell gave $\Delta\mathrm{C5}{=}{+}0.84$ for GPT-5.4 under F (H1-supported region); at $N{=}300$ the same effect collapses to $-0.12$ (Appendix~\ref{app:reversal}). This sign-reversing reversal motivates the $N{=}100$-per-condition-with-replicates sizing used throughout.

\textit{Channels are near-independent.} Pooled across 3{,}062 (case, condition, model) triples, between-channel correlations are low throughout. No pair exceeds the pre-registered $|\rho|>0.7$ flag threshold (Table~\ref{tab:correlations}). The strongest relationship is C2--C3 Spearman at $-0.29$, consistent with the construct distinction: larger edits mechanically touch more of the claim and can reduce scope overlap. The four channels are therefore empirically near-independent on the main data, supporting the \S\ref{sec:metric} design claim that they are jointly informative rather than redundant.
\begin{table}[t]
\centering
\caption{Channel correlations pooled across 3{,}062 (case, condition, model) triples. Pearson above the diagonal, Spearman below.}
\small
\begin{tabular*}{\columnwidth}{@{\extracolsep{\fill}}lcccc@{}}
\toprule
   & C1       & C2       & C3       & C5       \\
\midrule
C1 & ---      & $+0.04$  & $-0.21$  & $-0.05$  \\
C2 & $+0.07$  & ---      & $-0.14$  & $-0.04$  \\
C3 & $-0.17$  & $-0.29$  & ---      & $+0.06$  \\
C5 & $-0.01$  & $+0.03$  & $+0.04$  & ---      \\
\bottomrule
\end{tabular*}
\label{tab:correlations}
\end{table}

\textit{Directional consistency (C4).} The grand C4 of $0.56$ (Table~\ref{tab:c4-pairs}) splits into two kinds of near-chance cell. The retrieval-on-template cells (F,C5 $0.50$; G,C5 $0.47$) sit at chance \emph{because} they measure the prior-injection null; the manipulation-check cells (D,C5 $0.42$; Probe~C for C1 at $0.40$/$0.32$) are positive controls whose below-chance rates are a genuine tuning limitation, bearing on C1's sensitivity, while the FLAT predictions are well-calibrated (B,C1 $0.95$; B,C3 $0.64$). C4 is supplementary and carries none of the headline verdicts, which rest on the per-amendment channels C1--C3 and C5.

\subsection{Three commercial properties not confirmed}
\label{sec:commercial}

The empirical pattern in \S\ref{sec:taxonomy} does not support three commonly advertised properties of retrieval-augmented drafting, against pre-registered testing (H1 and H3 against fixed effect-size thresholds, H2 against a sign-and-magnitude equivalence condition).
\textbf{Retrieval grounds the output}: on grounding alignment (C1), $|\Delta\mathrm{C1}| \leq 0.014$ under both F and G across all four models, and no model shows a significant positive $\Delta\mathrm{C1}$ under either probe after Holm correction; the same null holds under a paraphrase-sensitive semantic C1 (Appendix~\ref{app:semantic-c1}). \textbf{Structural retrieval beats random retrieval}: H2 is rejected for three of four models; only Sonnet 4 satisfies $F \approx G$. \textbf{Retrieval makes models write more like experts}: the maximum observed $|\Delta\mathrm{C5}|$ is 0.20 (boundary on H1's threshold), and no model clears a symmetric $-0.20$ ``retrieval reduces template'' threshold either. These findings do not entail that retrieval-augmented patent drafting fails in general; they entail that the three properties are not confirmed in this instrument, and the released framework can be applied to other (model, corpus, retriever) configurations.

\subsection{Robustness of the null}
\label{sec:robustness}
The prior-injection null is stable along three axes we varied after the main run.
\textbf{Retrieval mechanism.} A dense (semantic) realization of Probe G, embedding-cosine top-3 over the same pool with $k$, exemplar format, and base prompt held fixed, retrieves far more semantically similar exemplars than either F or G (mean query--exemplar cosine 0.87 vs 0.59--0.61) yet moves nothing: across 1{,}200 calls (0 failures) on all four models, 0 of 16 baseline-versus-dense comparisons survive Holm correction and no model shows a significant C5 increase (Appendix~\ref{app:dense-g}). The null therefore spans three retrieval regimes (random, structural, dense), not just the deterministic matcher.
\textbf{Retrieval depth.} Varying only $k\in\{1,3,5,10\}$ with the cohort, retriever, scaffold, exemplar ordering, model, and decoding held fixed leaves C1 practically stable and yields no monotonic increase in C5; applying the pre-registered criteria at every depth gives the same qualitative verdicts, so the results are not an artifact of $k{=}3$ (Appendix~\ref{app:ksweep}).
\textbf{Grounding metric.} The grounding null holds under a paraphrase-sensitive semantic C1 that flags different cases than the lexical version (Spearman $\rho{=}0.37$; Appendix~\ref{app:semantic-c1}).

\section{Conclusion}
We presented a probe-based evaluation of rejection-grounded patent claim amendment: a corpus of 7,385 four-tuples from the USPTO Open Data Portal, seven probes, and a five-channel deterministic metric (C1--C3 and C5 carry the verdicts; C4 is supplementary). In this four-model, one-corpus setting, pre-registered testing finds no model above the H1 prior-injection threshold (max $\Delta$C5 $=+0.20$), a null that holds under random, structural, and dense retrieval and across $k\in\{1,3,5,10\}$ (\S\ref{sec:robustness}); H2 holds only for Sonnet 4, H3 only for Haiku 4.5, and revision locality (C2) reveals a model-specific difference (clearest for GPT-5.4) the template channel misses. We release the framework, corpus, and scoring code so the protocol can be applied to commercial stacks and to other retrieval configurations, where cell occupancy may differ.

\section*{\textit{Limitations}}
\paragraph{Scope and generalization.}
We examine a single rejection-amendment round; real prosecution can span multiple rounds with accumulating examiner commitments that our probes do not capture. Findings come from one domain (U.S. patent amendment) and four models; although we corroborate the null across random, structural, and dense retrieval and across retrieval depths $k\in\{1,3,5,10\}$ (\S\ref{sec:robustness}), generalization to other structured-document revision tasks such as legal briefs or scientific revision is conjectural and reserved for follow-up work. The retrieval pool is also drawn from the same corpus distribution as the test set, whereas a deployed commercial retriever would index a different corpus; we therefore read our retrieval effects as an upper bound on best-case benefit.

\paragraph{The empty prior-injector cell is conditional on the matcher.}
The headline finding that the prior-injector cell of the taxonomy is empty is conditional on Probe G's deterministic feature-based matcher (statute section $\times$ subsection $\times$ coarse limitation-pattern), chosen so that the F-versus-G contrast is interpretable as a structural-similarity test rather than a retriever-quality test (\S\ref{sec:ret-probes}, Appendix~\ref{app:retrieval}). A dense (semantic) realization of the same policy, however, leaves the null unchanged (Appendix~\ref{app:dense-g}), so the empty cell is not an artifact of the deterministic matcher; the null holds across random, structural, and dense retrieval. A still-different retriever indexing a different corpus could in principle relocate a model, and the released framework applies to such configurations without modification, so we treat the result as a controlled-policy data point rather than a closed claim about retrieval-augmented drafting in general.

\paragraph{Statistical adjudication.}
The 100-case cohort trades raw $n$ for representativeness and probe depth (\S\ref{sec:cohort}); further batches can be drawn at zero cost from the released sampling procedure (Appendix~\ref{app:replication}). We report paired Wilcoxon signed-rank tests with Holm--Bonferroni correction and bootstrap 95\% CIs on the headline $\Delta$C5 and $\Delta$C2 shifts, together with TOST equivalence tests against the pre-registered $\pm 0.2$ margin (Appendix~\ref{app:significance}). No $\Delta$C5 shift is significant after correction, and the median $\Delta$C5 is equivalent to zero within the pre-registered margin for all four models; the GPT-4o-mini boundary verdict is therefore supported by an equivalence result rather than only by adjacency to the threshold. Because per-case C5 varies widely ($\pm 2$ to $8$; \S\ref{sec:methodology-validity}), the heavy-tailed \emph{mean} $\Delta$C5 is not tightly bounded, so we frame the null as a statement about the central tendency and rank distribution rather than the mean. The sample-size reversal documented in \S\ref{sec:methodology-validity} motivates the $N{=}100$-with-replicates sizing, which we flag so that downstream users do not under-power their own runs.

\paragraph{Evaluation scope.}
A small fraction ($\sim$10\%) of cases fail parsing due to OCR degradation or missing XML in older file wrappers; results are reported stratified by amendment-size quartile (Appendix~\ref{app:parsing}) so that conclusions are conditional on the parseable subset rather than confounded by small-denominator cases. Prompts cap the verbatim examiner body at 8{,}000 characters, which binds in 80\% of cases (median body 13{,}448 characters); however, the structured rejection identification (attacked claim numbers, statute section and subsection, and cited references) precedes the body and is never truncated, so the material the metric scores against is retained in all 100 cases and a statutory-rejection statement remains within the retained window in every case. Per-model parse-success is 97.4\% overall (\S\ref{sec:results}); recomputing the H1 verdicts on the common subset of 65 cases parsed by all four models under every condition leaves the prior-injection null unchanged (maximum $\Delta$C5 $=+0.10$), and dropped cases do not differ substantially from the cohort in statute, amendment pattern, or technology center (Appendix~\ref{app:truncation}).
Finally, our channels are designed to track attorney-practice failure modes but are not a substitute for attorney judgment; a sample-level attorney-validation study (Appendix~\ref{app:attorney}) scores inter-rater agreement with the automated channels.

\section*{Ethical and legal considerations}
Patent claim amendment is a legally consequential act performed by registered agents. Our work evaluates whether LLMs could in principle perform this task grounded-ly; we take no position on whether they should be deployed for unassisted drafting. We release the corpus and probes to enable scrutiny of commercial claims, not to recommend replacement of human counsel. Data sources are public USPTO Open Data Portal artifacts with no copyright restriction. The corpus does not contain personal information beyond the inventor/attorney names that are part of the public record. We do not re-identify or link across filings beyond what ODP already exposes.

\paragraph{Use of AI assistants.}

We used commercial large language models as the subjects of evaluation (Claude Sonnet~4, Claude Haiku~4.5, GPT-5.4, GPT-4o-mini); their outputs constitute the experimental data scored by our metric. AI assistants were used during manuscript preparation for coding support, language and typo editing. It was not used for research ideation, experimental design, or the analysis itself.

\section*{Reproducibility Statement}
All artifacts required to reproduce the headline numbers are released under a permissive open license at \url{https://github.com/TeamLab/probing-rag-patent-amendment}:
\begin{itemize}\setlength\itemsep{2pt}
    \item the parsed JSONL corpus of 7{,}385 four-tuples and the parsing pipeline (\S\ref{corpus}, Appendix~\ref{app:corpus});
    \item the deterministic cohort-selection procedure with the batch~0 seed; additional replication batches are reproducible by re-running \texttt{select\_cohort.py} under any seed (Appendix~\ref{app:cohort}, Appendix~\ref{app:replication});
    \item the seven probe prompt templates and the five-channel metric implementation (Appendix~\ref{app:prompts}, Appendix~\ref{app:metric-code});
    \item the retrieval-pool index (Appendix~\ref{app:retrieval-pool});
    \item the runnable model-call driver, the five-channel scoring pipeline, and the analysis scripts that regenerate every table from raw model outputs, with pinned package versions (\texttt{requirements.txt}).
\end{itemize}
Reproducing the model calls requires API access to the four evaluated frontier LLMs; per-model cost estimates and inference settings are reported in Appendix~\ref{app:cost} and \S\ref{sec:model-matrix}.

\section*{Acknowledgments}
This research was supported by the ``Regional Growth and Talent Development System (Anchor)'' Project, funded by the Ministry of Education and Busan Metropolitan City (2026-Anchor-02-001-004, 20\%), and by the MSIT (Ministry of Science and ICT), Korea, under grants through the National Research Foundation of Korea (NRF) (No. RS-2024-00354675, 40\%) and the ICAN (ICT Challenge and Advanced Network of HRD) support program supervised by the IITP (Institute for Information \& Communications Technology Planning \& Evaluation) (IITP-2023-RS-2023-00259806, 20\%). This work was also partly supported by the Technology Development Program (TIPS, RS-2024-00554500, 10\%) funded by the Ministry of SMEs and Startups (MSS, Korea), and partly by the Korea Institute of Marine Science \& Technology Promotion (KIMST) funded by the Ministry of Oceans and Fisheries, Korea (RS-2026-25544055, 10\%).

\bibliography{custom}

\begin{thebibliography}{36}
\providecommand{\natexlab}[1]{#1}

\bibitem[{Asai et~al.(2024)Asai, Wu, Wang, Sil, and Hajishirzi}]{asai2023}
Akari Asai, Zeqiu Wu, Yizhong Wang, Avirup Sil, and Hannaneh Hajishirzi. 2024.
\newblock \href {https://arxiv.org/abs/2310.11511} {{Self-RAG}: Learning to
  retrieve, generate, and critique through self-reflection}.
\newblock In \emph{International Conference on Learning Representations}.

\bibitem[{Campbell and Fiske(1959)}]{campbell1959}
Donald~T. Campbell and Donald~W. Fiske. 1959.
\newblock \href {https://psycnet.apa.org/doi/10.1037/h0046016} {Convergent and
  discriminant validation by the multitrait-multimethod matrix}.
\newblock \emph{Psychological Bulletin}, 56(2):81--105.

\bibitem[{Chen et~al.(2021)Chen, Tworek, Jun, Yuan, de~Oliveira~Pinto, Kaplan,
  Edwards, Burda, Joseph, Brockman, Ray, Puri, Krueger, Petrov, Khlaaf, Sastry,
  Mishkin, Chan, Gray, Ryder, Pavlov, Power, Kaiser, Bavarian, Winter, Tillet,
  Such, Cummings, Plappert, Chantzis, Barnes, Herbert-Voss, Guss, Nichol,
  Paino, Tezak, Tang, Babuschkin, Balaji, Jain, Saunders, Hesse, Carr, Leike,
  Achiam, Misra, Morikawa, Radford, Knight, Brundage, Murati, Mayer, Welinder,
  McGrew, Amodei, McCandlish, Sutskever, and Zaremba}]{chen2021}
Mark Chen, Jerry Tworek, Heewoo Jun, Qiming Yuan, Henrique~Ponde
  de~Oliveira~Pinto, Jared Kaplan, Harri Edwards, Yuri Burda, Nicholas Joseph,
  Greg Brockman, Alex Ray, Raul Puri, Gretchen Krueger, Michael Petrov, Heidy
  Khlaaf, Girish Sastry, Pamela Mishkin, Brooke Chan, Scott Gray, and 39
  others. 2021.
\newblock \href {https://arxiv.org/abs/2107.03374} {Evaluating large language
  models trained on code}.
\newblock \emph{Preprint}, arXiv:2107.03374.

\bibitem[{Cuconasu et~al.(2024)Cuconasu, Trappolini, Siciliano, Filice,
  Campagnano, Maarek, Tonellotto, and Silvestri}]{cuconasu2024}
Florin Cuconasu, Giovanni Trappolini, Federico Siciliano, Simone Filice, Cesare
  Campagnano, Yoelle Maarek, Nicola Tonellotto, and Fabrizio Silvestri. 2024.
\newblock \href {https://arxiv.org/abs/2401.14887} {The power of noise:
  Redefining retrieval for {RAG} systems}.
\newblock In \emph{Proceedings of the 47th International ACM SIGIR Conference
  on Research and Development in Information Retrieval (SIGIR)}.

\bibitem[{Elazar et~al.(2021)Elazar, Ravfogel, Jacovi, and
  Goldberg}]{elazar2021}
Yanai Elazar, Shauli Ravfogel, Alon Jacovi, and Yoav Goldberg. 2021.
\newblock \href {https://arxiv.org/abs/2006.00995} {Amnesic probing: Behavioral
  explanation with amnesic counterfactuals}.
\newblock \emph{Transactions of the Association for Computational Linguistics},
  9:160--175.

\bibitem[{Es et~al.(2024)Es, James, Espinosa-Anke, and Schockaert}]{es2024}
Shahul Es, Jithin James, Luis Espinosa-Anke, and Steven Schockaert. 2024.
\newblock \href {https://arxiv.org/abs/2309.15217} {{RAGAS}: Automated
  evaluation of retrieval augmented generation}.
\newblock In \emph{Proceedings of EACL (System Demonstrations)}.

\bibitem[{Fang et~al.(2024)Fang, Bai, Ni, Yang, Chen, and Xu}]{fang2024raat}
Feiteng Fang, Yuelin Bai, Shiwen Ni, Min Yang, Xiaojun Chen, and Ruifeng Xu.
  2024.
\newblock \href {https://arxiv.org/abs/2405.20978} {Enhancing noise robustness
  of retrieval-augmented language models with adaptive adversarial training}.
\newblock In \emph{Proceedings of the 62nd Annual Meeting of the Association
  for Computational Linguistics (ACL)}.

\bibitem[{Gottweis et~al.(2026)Gottweis, Weng, Daryin, Tu, Palepu, Sirkovic,
  Myaskovsky, Weissenberger, Rong, Tanno, Saab, Popovici, Blum, Zhang, Chou,
  Hassidim, Gokturk, Vahdat, Kohli, Matias, Carroll, Kulkarni, Tomasev, Guan,
  Dhillon, Vaishnav, Lee, Costa, Penad{\'e}s, Peltz, Xu, Pawlosky,
  Karthikesalingam, and Natarajan}]{gottweis2025}
Juraj Gottweis, Wei-Hung Weng, Alexander Daryin, Tao Tu, Anil Palepu, Petar
  Sirkovic, Artiom Myaskovsky, Felix Weissenberger, Keran Rong, Ryutaro Tanno,
  Khaled Saab, Dan Popovici, Jacob Blum, Fan Zhang, Katherine Chou, Avinatan
  Hassidim, Burak Gokturk, Amin Vahdat, Pushmeet Kohli, and 15 others. 2026.
\newblock \href {https://www.nature.com/articles/s41586-026-10644-y}
  {Accelerating scientific discovery with {Co-Scientist}}.
\newblock \emph{Nature}.

\bibitem[{Guu et~al.(2020)Guu, Lee, Tung, Pasupat, and Chang}]{guu2020}
Kelvin Guu, Kenton Lee, Zora Tung, Panupong Pasupat, and Ming-Wei Chang. 2020.
\newblock \href {https://arxiv.org/abs/2002.08909} {{REALM}:
  Retrieval-augmented language model pre-training}.
\newblock In \emph{International Conference on Machine Learning}.

\bibitem[{Izacard et~al.(2023)Izacard, Lewis, Lomeli, Hosseini, Petroni,
  Schick, Dwivedi-Yu, Joulin, Riedel, and Grave}]{izacard2022}
Gautier Izacard, Patrick Lewis, Maria Lomeli, Lucas Hosseini, Fabio Petroni,
  Timo Schick, Jane Dwivedi-Yu, Armand Joulin, Sebastian Riedel, and Edouard
  Grave. 2023.
\newblock \href {https://www.jmlr.org/papers/v24/23-0037.html} {{Atlas}:
  Few-shot learning with retrieval augmented language models}.
\newblock \emph{Journal of Machine Learning Research}.
\newblock ArXiv:2208.03299 (2022).

\bibitem[{Jang et~al.(2025)Jang, Lee, Min, and Choi}]{jang2025}
Yehoon Jang, Chaewon Lee, Hyun-seok Min, and Sungchul Choi. 2025.
\newblock \href {https://aclanthology.org/2025.nllp-1.17/} {{PILOT}-bench: A
  benchmark for legal reasoning in the patent domain with {IRAC}-aligned
  classification tasks}.
\newblock In \emph{Proceedings of the Natural Legal Language Processing
  Workshop (NLLP), at EMNLP}.

\bibitem[{Jiang et~al.(2025{\natexlab{a}})Jiang, Scherz, and Goetz}]{jiang2024}
Lekang Jiang, Pascal~A. Scherz, and Stephan Goetz. 2025{\natexlab{a}}.
\newblock \href {https://aclanthology.org/2025.naacl-long.116/} {{Patent-CR}: A
  dataset for patent claim revision}.
\newblock In \emph{Proceedings of NAACL (Long Papers)}.

\bibitem[{Jiang et~al.(2025{\natexlab{b}})Jiang, Scherz, and Goetz}]{jiang2025}
Lekang Jiang, Pascal~A. Scherz, and Stephan Goetz. 2025{\natexlab{b}}.
\newblock \href {https://aclanthology.org/2025.acl-long.190/} {Towards better
  evaluation for generated patent claims}.
\newblock In \emph{Proceedings of ACL (Long Papers)}, pages 3775--3788.

\bibitem[{Jimenez et~al.(2024)Jimenez, Yang, Wettig, Yao, Pei, Press, and
  Narasimhan}]{jimenez2024}
Carlos~E. Jimenez, John Yang, Alexander Wettig, Shunyu Yao, Kexin Pei, Ofir
  Press, and Karthik Narasimhan. 2024.
\newblock \href {https://arxiv.org/abs/2310.06770} {{SWE}-bench: Can language
  models resolve real-world {GitHub} issues?}
\newblock In \emph{International Conference on Learning Representations}.

\bibitem[{Kawano et~al.(2024)Kawano, Nonaka, and Yoshino}]{claimbrush2024}
Seiya Kawano, Hirofumi Nonaka, and Koichiro Yoshino. 2024.
\newblock \href {https://arxiv.org/abs/2410.05575} {{ClaimBrush}: A novel
  framework for automated patent claim refinement based on large language
  models}.
\newblock In \emph{IEEE International Conference on Big Data (BigData)}.

\bibitem[{Knappich et~al.(2025)Knappich, Friedrich, H{\"a}tty, and
  Razniewski}]{knappich2025}
Valentin Knappich, Annemarie Friedrich, Anna H{\"a}tty, and Simon Razniewski.
  2025.
\newblock \href {https://ceur-ws.org/Vol-4062/paper3.pdf} {{PEDANTIC}: A
  dataset for the automatic examination of definiteness in patent claims}.
\newblock In \emph{Proceedings of PatentSemTech}.

\bibitem[{Lee(2023)}]{lee2023}
Jieh-Sheng Lee. 2023.
\newblock \href
  {https://www.sciencedirect.com/science/article/pii/S0172219023000030}
  {Evaluating generative patent language models}.
\newblock \emph{World Patent Information}.

\bibitem[{Lee(2024)}]{instructpatentgpt2024}
Jieh-Sheng Lee. 2024.
\newblock \href {https://arxiv.org/abs/2406.16897} {{InstructPatentGPT}:
  Training patent language models to follow instructions with human feedback}.
\newblock \emph{Preprint}, arXiv:2406.16897.

\bibitem[{Lewis et~al.(2020)Lewis, Perez, Piktus, Petroni, Karpukhin, Goyal,
  K{\"u}ttler, Lewis, Yih, Rockt{\"a}schel, Riedel, and Kiela}]{lewis2020}
Patrick Lewis, Ethan Perez, Aleksandra Piktus, Fabio Petroni, Vladimir
  Karpukhin, Naman Goyal, Heinrich K{\"u}ttler, Mike Lewis, Wen-tau Yih, Tim
  Rockt{\"a}schel, Sebastian Riedel, and Douwe Kiela. 2020.
\newblock \href {https://arxiv.org/abs/2005.11401} {Retrieval-augmented
  generation for knowledge-intensive {NLP} tasks}.
\newblock In \emph{Advances in Neural Information Processing Systems}.

\bibitem[{Lim et~al.(2025)Lim, Nam, Na, Cho, Yang, Shin, Lee, Kim, Lee, and
  Hong}]{lim2025}
Hyunseung Lim, Sooyohn Nam, Sungmin Na, Ji~Yong Cho, June~Yong Yang, Hyungyu
  Shin, Yoonjoo Lee, Juho Kim, Moontae Lee, and Hwajung Hong. 2025.
\newblock \href {https://neurips.cc/virtual/2025/poster/121720} {{PANORAMA}: A
  dataset and benchmarks capturing decision trails and rationales in patent
  examination}.
\newblock In \emph{Advances in Neural Information Processing Systems
  (NeurIPS)}.

\bibitem[{Liu et~al.(2024)Liu, Lin, Hewitt, Paranjape, Bevilacqua, Petroni, and
  Liang}]{liu2024lost}
Nelson~F. Liu, Kevin Lin, John Hewitt, Ashwin Paranjape, Michele Bevilacqua,
  Fabio Petroni, and Percy Liang. 2024.
\newblock Lost in the middle: How language models use long contexts.
\newblock \emph{Transactions of the Association for Computational Linguistics},
  12.

\bibitem[{Lu et~al.(2026)Lu, Lu, Tjarko~Lange, Yamada, Hu, Foerster, Ha, and
  Clune}]{lu2026}
Chris Lu, Cong Lu, Robert Tjarko~Lange, Yutaro Yamada, Shengran Hu, Jakob
  Foerster, David Ha, and Jeff Clune. 2026.
\newblock \href {https://www.nature.com/articles/s41586-026-10265-5} {Towards
  end-to-end automation of {AI} research}.
\newblock \emph{Nature}.

\bibitem[{Ribeiro et~al.(2020)Ribeiro, Wu, Guestrin, and Singh}]{ribeiro2020}
Marco~T{\'u}lio Ribeiro, Tongshuang Wu, Carlos Guestrin, and Sameer Singh.
  2020.
\newblock \href {https://arxiv.org/abs/2005.04118} {Beyond accuracy: Behavioral
  testing of {NLP} models with {CheckList}}.
\newblock In \emph{Proceedings of ACL}.
\newblock ACL 2020 Best Paper.

\bibitem[{Sharma et~al.(2019)Sharma, Li, and Wang}]{sharma2019}
Eva Sharma, Chen Li, and Lu~Wang. 2019.
\newblock \href {https://arxiv.org/abs/1906.03741} {{BIGPATENT}: A large-scale
  dataset for abstractive and coherent summarization}.
\newblock In \emph{Proceedings of ACL}.

\bibitem[{Shi et~al.(2025)Shi, Li, Zhang, Fang, Wang, Liu, Zhao, Zhu, Gao,
  Zhong, Zhang, Ke, E, Cai, and Wang}]{shi2024}
Yaorui Shi, Sihang Li, Taiyan Zhang, Xi~Fang, Jiankun Wang, Zhiyuan Liu,
  Guojiang Zhao, Zhengdan Zhu, Zhifeng Gao, Renxin Zhong, Linfeng Zhang, Guolin
  Ke, Weinan E, Hengxing Cai, and Xiang Wang. 2025.
\newblock \href {https://arxiv.org/abs/2412.07819} {Intelligent system for
  automated molecular patent infringement assessment}.
\newblock \emph{Preprint}, arXiv:2412.07819.

\bibitem[{Shomee et~al.(2025)Shomee, Maity, and Medya}]{shomee2025}
Homaira~Huda Shomee, Suman~Kalyan Maity, and Sourav Medya. 2025.
\newblock \href {https://arxiv.org/abs/2507.22387} {{PATENTWRITER}: A
  benchmarking study for patent drafting with {LLMs}}.
\newblock \emph{Preprint}, arXiv:2507.22387.

\bibitem[{Song et~al.(2026)Song, Song, Pfister, and Yoon}]{song2026}
Yiwen Song, Yale Song, Tomas Pfister, and Jinsung Yoon. 2026.
\newblock \href {https://arxiv.org/abs/2604.05018} {{PaperOrchestra}: A
  multi-agent framework for automated {AI} research paper writing}.
\newblock \emph{Preprint}, arXiv:2604.05018.

\bibitem[{Thakur et~al.(2021)Thakur, Reimers, R{\"u}ckl{\'e}, Srivastava, and
  Gurevych}]{thakur2021}
Nandan Thakur, Nils Reimers, Andreas R{\"u}ckl{\'e}, Abhishek Srivastava, and
  Iryna Gurevych. 2021.
\newblock \href {https://arxiv.org/abs/2104.08663} {{BEIR}: A heterogeneous
  benchmark for zero-shot evaluation of information retrieval models}.
\newblock In \emph{NeurIPS Datasets and Benchmarks Track}.

\bibitem[{{U.S. Patent and Trademark Office}(2025)}]{uspto2025}
{U.S. Patent and Trademark Office}. 2025.
\newblock Patents dashboard.
\newblock \url{https://www.uspto.gov/dashboard/patents/}.

\bibitem[{{U.S. Patent and Trademark Office, Office of the Chief
  Economist}(2017)}]{oard2017}
{U.S. Patent and Trademark Office, Office of the Chief Economist}. 2017.
\newblock Office action research dataset for patents.
\newblock USPTO Open Data.
\newblock
  \url{https://www.uspto.gov/ip-policy/economic-research/research-datasets}.

\bibitem[{van Miltenburg et~al.(2021)van Miltenburg, van~der Lee, and
  Krahmer}]{vanmiltenburg2021}
Emiel van Miltenburg, Chris van~der Lee, and Emiel Krahmer. 2021.
\newblock \href {https://aclanthology.org/2021.naacl-main.51/} {Preregistering
  {NLP} research}.
\newblock In \emph{Proceedings of NAACL}.
\newblock NAACL 2021 Best Thematic Paper.

\bibitem[{Van~Noorden(2014)}]{vannoorden2014}
Richard Van~Noorden. 2014.
\newblock \href {https://www.nature.com/articles/nature.2014.14763} {Publishers
  withdraw more than 120 gibberish papers}.
\newblock \emph{Nature News}.

\bibitem[{Vig et~al.(2020)Vig, Gehrmann, Belinkov, Qian, Nevo, Singer, and
  Shieber}]{vig2020}
Jesse Vig, Sebastian Gehrmann, Yonatan Belinkov, Sharon Qian, Daniel Nevo,
  Yaron Singer, and Stuart Shieber. 2020.
\newblock \href {https://arxiv.org/abs/2004.12265} {Investigating gender bias
  in language models using causal mediation analysis}.
\newblock In \emph{Advances in Neural Information Processing Systems}.

\bibitem[{Wang et~al.(2024)Wang, Ni, Liu, Chen, Feng, Wei, Qu, Alinejad-Rokny,
  Lin, and Yang}]{wang2024}
Qiyao Wang, Shiwen Ni, Huaren Liu, Guhong Chen, Xi~Feng, Chi Wei, Qiang Qu,
  Hamid Alinejad-Rokny, Yuan Lin, and Min Yang. 2024.
\newblock \href {https://arxiv.org/abs/2412.09796} {{AutoPatent}: A multi-agent
  framework for automatic patent generation}.
\newblock \emph{Preprint}, arXiv:2412.09796.

\bibitem[{Yoo et~al.(2025)Yoo, Xu, and Cao}]{yoo2025}
Yongmin Yoo, Qiongkai Xu, and Longbing Cao. 2025.
\newblock \href {https://aclanthology.org/2025.emnlp-main.1564/}
  {{PatentScore}: Multi-dimensional evaluation of {LLM}-generated patent
  claims}.
\newblock In \emph{Proceedings of EMNLP}.

\bibitem[{Yu et~al.(2025)Yu, Liang, and Hu}]{treeofclaims2025}
Shuyang Yu, Jianan Liang, and Hui Hu. 2025.
\newblock \href {https://arxiv.org/abs/2511.16972} {{ToC}: Tree-of-claims
  search with multi-agent language models}.
\newblock \emph{Preprint}, arXiv:2511.16972.

\end{thebibliography}
\bibliographystyle{acl_natbib}
\clearpage
\appendix
\section{Corpus construction pipeline}
\label{app:corpus}

\subsection{Source pool and enrichment chain}
\label{app:source}

We start from the PILOT-Bench \citep{jang2025} PTAB-appeal subset of 13,749 proceedings, each carrying a patent application number, decision outcome, statute citations, and technology-center assignment. For each proceeding we run a five-stage enrichment against the USPTO Open Data Portal (ODP): (i) resolve the application number via ODP query and obtain the file-wrapper document list; (ii) identify the first non-final rejection (USPTO code CTNF, ``correspondence: notice of non-final rejection''), the immediately preceding incoming Claims filing (pre-amendment), and the first incoming Claims filing after the CTNF (post-amendment); (iii) download each as XML, excluding filings available only as PDF/image scans; (iv) parse XML under three schema families (\S\ref{app:parsing}); (v) compute per-case diff with status $\in$ \{kept, modified, new, cancelled\} via claim-number alignment, plus SequenceMatcher similarity for modified claims.

\subsection{XML schema handling}
\label{app:parsing}

Three schema families coexist in ODP's claim documents:
\begin{itemize}\setlength\itemsep{2pt}
    \item \textbf{Legacy DTD} \texttt{<us-patent-application>} for pre-2014 filings. Claim text is inline; amendments use \texttt{[[deleted]]} bracket markup. Our parser splits claim bodies on numeric headers and extracts bracket spans.
    \item \textbf{USPTO ClaimsDocument v1.3} \texttt{<pat:ClaimsDocument>} for mid-generation filings. Uses \texttt{<pat:Claim><pat:ClaimText>} wrappers with underline and \texttt{<pat:DeletedText>} markup.
    \item \textbf{ST96 v2} \texttt{<uspat:ClaimsDocument>} for recent filings. Dual URI namespace; \texttt{<pat:Ins>}/\texttt{<pat:Del>} insertion and deletion tags; \texttt{<ImplicitClaim>} elements encode claim-number ranges that expand into separate records.
\end{itemize}

Rejection documents use \texttt{<uspat:OutgoingDocument>} with structured \texttt{<uscom:FormParagraph>} and \texttt{<uscom:DataField>} slots encoding claim numbers, statute subsection, rejection type, and cited references. V7.1 documents replace semantic FormParagraphNumber identifiers with generic IDs; for those we fall back to regular-expression parsing of the prose body. OCR-confidence wrappers (\texttt{<pat:OCRConfidenceData>} and the legacy \texttt{<confidence>}) are stripped before parsing.

\subsection{Parser hardening}
\label{parser-hardening}
A targeted hardening pass on 2026-04-18 raised cohort coverage from 76/100 to 100/100 via four sequential fixes: dual-URI namespace fallback plus \texttt{<ImplicitClaim>} range expansion (93/100); inline-header fallback with paren-form and multi-inline-text wrapper splits (97/100); bracket \texttt{[Claim N]} and word-prefix \texttt{Claim N (Status):} matching with DTD legacy inline fallback (99/100); and a deterministic case replacement (Appendix~\ref{app:cohort}). The same hardening applied to the full $\beta$ corpus recovered several thousand additional cases that the baseline parser had silently dropped.

\subsection{Final corpus statistics}
\label{app:funnel}

Funnel from PILOT-Bench source pool to release corpus and cohort eligibility:

\begin{itemize}\setlength\itemsep{2pt}
    \item PILOT-Bench PTAB-appeal subset: 13,749 proceedings.
    \item ODP retrieval targets (resolvable application numbers): 9,956.
    \item Parseable (pre, post, CTNF) triples: 7,385 (74.2\% parse yield).
    \item Case-level C2 computable: 5,755.
    \item Cohort eligibility pool (after parse-health, axis coverage, and data-defect filters): 4,270.
\end{itemize}
The remaining 2,571 parse failures concentrate in documents without XML download options (older filings), publications not found in ODP, and a small set of malformed wrappers that survive parser hardening. Across the full 7,385-case corpus: per-claim status modified 99{,}335 (64.5\%), cancelled 33{,}118 (21.5\%), new 17{,}008 (11.0\%), kept 4{,}459 (2.9\%). Modified-claim similarity ratio median 0.933, mean 0.841, Q1 0.827, Q3 0.966 ($n = 99{,}335$). Case-level pre-to-post Levenshtein similarity median 0.72, Q1 0.55, Q3 0.84 ($n = 5{,}755$). First-rejection statute distribution: \S102 2{,}337 (31.6\%), \S112 1{,}245 (16.9\%), \S101 950 (12.9\%), \S103 738 (10.0\%). XML format mix: DTD legacy 3{,}576 (48.4\%), namespaced (v1.3 plus ST96) 3{,}808 (51.6\%). Coarse amendment pattern: modify\_only 5{,}043 (68.3\%), cancel\_heavy 1{,}410 (19.1\%), other 757 (10.3\%), new\_or\_add 175 (2.4\%).

\subsection{Relation to other corpora}
\label{app:relation}
The corpus is constructed to enable grounded-revision evaluation (pre $\leftrightarrow$ rejection $\leftrightarrow$ prior art $\leftrightarrow$ post) rather than to maximize raw scale. As noted in \S\ref{sec:corpus-release}, ours is the only publicly described corpus that aligns rejection context, cited prior art, pre/post claim pair, and amendment diff at the XML level. Patent-CR provides the pre/post pair only; PANORAMA, PEDANTIC, and PILOT-Bench target judgment, classification, or upstream retrieval rather than applicant-side amendment.

\section{Per-model cost estimates}
\label{app:cost}
Per-call token counts and vendor pricing yield approximately \$110 for Stage 1 (Claude Sonnet 4 and GPT-5.4 across the full probe battery on 100 cases $\times$ 8 conditions $\times$ 3 replicates) and approximately \$110 for Stage 2 (Claude Haiku 4.5 and GPT-4o-mini under the same protocol), for a combined experimental cost of approximately \$220 across 9{,}600 model calls. Infrastructure cost is zero: USPTO ODP is a public API and all compute is local.

\section{Worked examples per channel}
\label{app:construct-checks}
This appendix maps each channel's numeric output to a concrete within-case contrast drawn from the main experiment. Scores are illustrative; they show what each channel responds to, not where population medians sit.

\subsection{C1 Grounding alignment}
\textbf{Case.} PILOT-2019005776 (Reversed, \S101-related, TC 2100, 2020--25, cancel\_heavy). \textbf{Model.} GPT-4o-mini. Under baseline the model's modified limitations overlapped the rejected limitations at rate $0.014$ (edit off-target). Under Probe A (Claim Truncation) the overlap rose to $0.402$ within-case, an order-of-magnitude shift on the same case and model. C1 catches whether an amendment lands on the examiner-attacked limitation rather than elsewhere in the claim.

\subsection{C2 Revision Locality}
\textbf{Case.} PILOT-2020002555 (Reversed, \S103+112, TC other, 2020--25, cancel\_heavy). \textbf{Model.} GPT-4o-mini. The gold amendment has edit distance 591 characters. Under baseline C2 $= 0.26$ (26\% of gold scale, an under-edit). Under Probe F (Random Retrieval) C2 $= 7.40$ (over $7\times$ gold scale, a near-complete rewrite). The channel registers the two misses as categorically different rather than as a single failure mode.

\subsection{C3 Scope preservation}
\textbf{Case.} PILOT-2020003998 (AIP, \S103+112, TC 3700, 2020--25, cancel\_heavy). \textbf{Model.} Claude Haiku 4.5. Under baseline noun-phrase Jaccard with the invention core is $0.983$. Under Probe D (Boilerplate Injection) it collapses to $0.073$ as the injected canonical phrases overwrite the invention-core noun phrases. This is the over-narrowing / redirection failure mode C3 targets.

\subsection{C5 Template dependence}
\textbf{Case.} PILOT-2013007708 (Affirmed, \S103, TC 3700, 2010--14, modify\_only). \textbf{Model.} GPT-4o-mini. Under baseline C5 $= 3.34$ canonical-phrase hits per 1{,}000 characters (7 hits in 2{,}081 characters). Under Probe G (Structural Retrieval) C5 $= 11.66$ (49 hits in 4{,}181 characters), a $3.5\times$ increase. This is the case $\gamma$ referenced in Appendix~\ref{app:reversal}.

\subsection{Simpler-baseline comparison}
\label{app:simpler-baselines}
On each worked example, BLEU/ROUGE/BERTScore baselines fail to register the within-case direction our channels detect. On PILOT-2020002555, BLEU-4 scores baseline and F as equally distant from gold ($\sim$0.12 vs $\sim$0.10), while C2 reads them as $0.26$ vs $7.40$. On PILOT-2020003998, BERTScore-F1 moves from 0.91 to 0.73 under Probe D, but the magnitude does not flag the invention-core-overwrite failure mode; C3 moves $0.98$ to $0.07$ and registers the scope collapse directly. On PILOT-2013007708, raw n-gram recurrence rises $1.2\times$ baseline (domain-natural repetition) while C5 rises $3.5\times$, isolating retrieval-induced template reuse from intrinsic patent prose repetition.

\subsection{Case $\beta$: GPT-5.4 F-vs-G sign flip}
\label{app:case-studies}

\textbf{Case.} PILOT-2020006264 (Affirmed, \S101-related, TC 3600, modify\_only). \textbf{Model.} GPT-5.4. C5 baseline 11.3 moves to 13.5 under F ($\Delta_F = +2.25$, inflation) and 9.8 under G ($\Delta_G = -1.50$, suppression). Same input, same retrieval pool, same $k$; only the selection rule differs, yet the effect inverts in sign. The other three models on this case: Sonnet 4 near-flat ($\Delta_F = +0.01$, $\Delta_G = -0.14$); Haiku 4.5 mildly inflates on F ($+1.96$) but is flat on G ($-0.29$); GPT-4o-mini consistently reduces ($\Delta_F = -3.20$, $\Delta_G = -3.32$). This case concretely realizes the H2-rejection verdict in \S\ref{sec:verdicts}: for GPT-5.4, random versus structural retrieval changes not just the magnitude but the sign of the template-dependence effect.

\section{Sample-size reversal}
\label{app:reversal}

The Table~\ref{tab:verdicts} verdicts are computed at $N{=}300$ per cell (100 cases $\times$ 3 replicates). A preliminary smoke run at $N{=}10$ per cell, on identical inputs and the same scoring code, gave $\Delta\mathrm{C5}{=}+0.84$ for GPT-5.4 under F, a magnitude that would have placed GPT-5.4 firmly in the H1-supported cell. At $N{=}300$ the same effect collapses to $-0.12$, a complete sign reversal. The reversal is sampling variability on a 100-case probe-based cohort used at $N{=}10$, not a calibration issue, and is consistent with the per-case dispersion of $\pm 2$ to $8$ that \S\ref{sec:verdicts} reports on this channel.

We document the GPT-5.4/F cell explicitly because it was the cell that triggered the $N{=}10 \rightarrow N{=}300$ escalation decision and is the largest documented reversal in our pipeline; the smoke run did not include a systematic $N{=}10$ replication on every $(\text{model}, \text{condition})$ cell in Table~\ref{tab:verdicts}, so we report this one reversal as a benchmark-methodology finding rather than as a cell-by-cell stability sweep. The implication for downstream users is the same in either case: replications should be sized at or above the pre-registered $N{=}100$ per condition with replicates rather than relying on smoke-run effect sizes, because at $N{=}10$ a sign-reversing reversal is plausible on any single cell of a 100-case probe-based cohort.

\section{Probe-D anchor: per-model breakdown}
\label{app:probe-d-anchor}

Extending the cross-model comparison in \S\ref{sec:taxonomy}, the per-model $|\Delta_F|$-vs-$|\Delta_D|$ gaps on three of four models are within $0.13$ (the maximum observed gap on those three models, reported descriptively as a post-hoc magnitude rather than as a pre-registered tolerance), indicating that random retrieval does not behave qualitatively differently from canonical boilerplate on Sonnet 4, GPT-5.4, and GPT-4o-mini. Haiku 4.5 is the per-model exception: its $\Delta_F = +0.19$ exceeds $|\Delta_D| = 0.09$ in absolute magnitude, consistent with its cell-(d) ``inert / mild prior-injector'' assignment, where retrieval produces the largest random-F template inflation in the panel (though still below the $+0.2$ H1 threshold). Even on Haiku 4.5, the retrieval shift is the same order as the boilerplate-prepend dosage. The cross-model qualitative reading therefore still holds: random retrieval is bounded by a generic prepended-context effect, with per-model magnitudes reported individually rather than averaged.

\section{Probe x channel full table}
\label{app:retrieval}

Medians across 100 cohort cases $\times$ 3 replicates per (model, condition). Channels C1, C2, C3, and C5 are shown; C4 is a directional-hit-rate measure with paired-probe structure, reported separately in Appendix~\ref{app:c4}. Probe G uses a deterministic feature-based matcher rather than a dense retriever so that the F vs G contrast is interpretable as a structural-similarity test rather than a retriever-quality test. An embedding retriever would conflate the structural-similarity signal we want to isolate with the representation-learning quirks of whichever embedder is chosen.

\begin{table}[t]
\centering
\caption{Per-model probe$\times$channel deltas. Conditions A--E are observational; F and G are retrieval interventions. Leading zeros omitted in baselines for compactness.}
\label{tab:probe-deltas}
\footnotesize
\setlength{\tabcolsep}{4pt}
\begin{tabular}{lcccc}
\toprule
cond & $\Delta$C1 & $\Delta$C2 & $\Delta$C3 & $\Delta$C5 \\
\midrule
\multicolumn{5}{l}{\textit{Sonnet 4} (base: C1 .009, C2 .617, C3 .887, C5 3.99)} \\
A & $+0.004$ & $-0.083$ & $-0.005$ & $+0.01$ \\
B & $+0.001$ & $-0.038$ & $-0.018$ & $+0.08$ \\
C & $-0.001$ & $+0.001$ & $-0.006$ & $-0.09$ \\
D & $+0.001$ & $+0.047$ & $-0.026$ & $+0.03$ \\
E & $+0.001$ & $-0.088$ & $-0.002$ & $-0.30$ \\
F & $-0.001$ & $-0.074$ & $-0.004$ & $-0.16$ \\
G & $-0.001$ & $-0.013$ & $-0.007$ & $-0.13$ \\
\midrule
\multicolumn{5}{l}{\textit{Haiku 4.5} (base: C1 .016, C2 1.147, C3 .738, C5 3.40)} \\
A & $+0.013$ & $-0.090$ & $+0.001$ & $+0.16$ \\
B & $-0.001$ & $-0.035$ & $-0.004$ & $+0.16$ \\
C & $+0.003$ & $-0.103$ & $-0.007$ & $+0.03$ \\
D & $-0.000$ & $+0.032$ & $-0.017$ & $-0.09$ \\
E & $-0.003$ & $-0.060$ & $+0.002$ & $+0.16$ \\
F & $-0.001$ & $-0.174$ & $+0.027$ & $+0.19$ \\
G & $-0.002$ & $-0.189$ & $+0.038$ & $+0.03$ \\
\midrule
\multicolumn{5}{l}{\textit{GPT-5.4} (base: C1 .008, C2 .789, C3 .892, C5 3.73)} \\
A & $+0.005$ & $-0.035$ & $+0.003$ & $+0.19$ \\
B & $+0.001$ & $-0.087$ & $+0.003$ & $+0.06$ \\
C & $-0.000$ & $-0.038$ & $+0.009$ & $+0.10$ \\
D & $-0.000$ & $+0.051$ & $-0.003$ & $+0.07$ \\
E & $-0.002$ & $-0.099$ & $+0.020$ & $+0.06$ \\
F & $+0.000$ & $+0.016$ & $-0.001$ & $-0.12$ \\
G & $+0.002$ & $+0.058$ & $+0.002$ & $+0.06$ \\
\midrule
\multicolumn{5}{l}{\textit{GPT-4o-mini} (base: C1 .017, C2 1.459, C3 .798, C5 3.14)} \\
A & $+0.018$ & $-0.035$ & $-0.007$ & $+0.02$ \\
B & $-0.009$ & $-0.050$ & $-0.030$ & $+0.08$ \\
C & $-0.001$ & $-0.047$ & $+0.023$ & $-0.29$ \\
D & $-0.009$ & $-0.044$ & $+0.034$ & $-0.07$ \\
E & $-0.007$ & $-0.008$ & $+0.028$ & $-0.07$ \\
F & $-0.006$ & $-0.044$ & $+0.021$ & $+0.20$ \\
G & $-0.006$ & $-0.076$ & $+0.028$ & $-0.27$ \\
\bottomrule
\end{tabular}
\end{table}

\section{C4 robustness per probe-pair}
\label{app:c4}

Directional hit-rate per (probe, channel) pair, pooled across the four models (Claude Sonnet~4, Claude Haiku~4.5, GPT-5.4, GPT-4o-mini) and three replicates on the 100-case cohort. Each case-model contrast scores 1 if the per-case $\Delta$ from baseline matches the pre-registered direction (UP / DOWN / FLAT with $|\Delta|<0.05$ threshold), 0 otherwise. The grand C4 in Table~\ref{tab:c4-pairs} is the unweighted mean over all hits, $0.56$ on $n{=}2{,}276$ case-model contrasts. C subconditions ($C_{\text{decoy}}$, $C_{\text{mechtrue}}$) are split per the probe-C index (Appendix~\ref{app:retrieval-pool}); 10 of 100 cohort cases are skipped at probe-build time, so the C-row $n$ is smaller than the others.

\begin{table}[t]
\centering
\caption{C4 directional hit-rate per (probe, channel) pair, pooled across four models and three replicates on the 100-case cohort. FLAT predictions use a $|\Delta|<0.05$ tolerance.}
\label{tab:c4-pairs}
\small
\begin{tabular*}{\columnwidth}{@{\extracolsep{\fill}}lcrr@{}}
\toprule
(probe, channel)          & predicted & hit-rate & $n$ \\
\midrule
B, C1                     & FLAT  & $0.95$ & $382$ \\
B, C3                     & FLAT  & $0.64$ & $387$ \\
F, C5                     & UP    & $0.50$ & $383$ \\
G, C5                     & UP    & $0.47$ & $384$ \\
D, C5                     & UP    & $0.42$ & $392$ \\
$C_{\text{mechtrue}}$, C1 & UP    & $0.40$ & $178$ \\
$C_{\text{decoy}}$, C1    & DOWN  & $0.32$ & $170$ \\
\midrule
grand C4                  &       & $0.56$ & $2{,}276$ \\
\bottomrule
\end{tabular*}
\end{table}

\section{Probe prompt templates}
\label{app:prompts}

The shared system prompt and base user-prompt template are released in \texttt{contexts/system.txt} and \texttt{contexts/user\_base.txt}. Probe-specific perturbations (A--G) are applied by \texttt{scripts/build\_probe\_prompts.py}: the Probe D boilerplate block, the Probe E drafting hints, and the Probe F/G retrieval-insertion templates are defined inline in that script.

\section{Five-channel metric computation}
\label{app:metric-code}

Reference implementations for channels C1 through C5 are provided in the supplementary material (\texttt{scripts/compute\_c\{1,2,3,4,5\}.py}). Each channel is a deterministic function of generated text, parser output, and the XML gold post-amendment claim, as defined in \S\ref{sec:channels}.

\section{Attorney validation study}
\label{app:attorney}

A 30--50-case double-rated sample assessing inter-rater agreement between our automated channels and licensed patent attorneys is committed post-review. Reported metrics will include Cohen's $\kappa$ for categorical channels and Spearman $\rho$ for continuous channels, against the automated scoring on the same cases. A 10-case pre-release sanity sample showed all four channels moving in the direction the reading attorney predicted; the full $\kappa$ quantification is reserved for the published study.

The 10 cases used in the sanity sample are the same 10 cases across all four channels (a single shared subset, not a per-channel subset), drawn deterministically from the 100-case cohort by application-number ordering, so that each case contributes one C1, C2, C3, and C5 reading to the attorney's directional check. Application-number ordering correlates loosely with filing date and therefore with XML-format era, so the 10-case sanity sample is not stratum-balanced; the full 30--50-case study will instead draw a stratum-balanced sample across the six cohort axes so that the directional check is not era-biased. The reading attorney was blind to the automated scores at the time of directional prediction: per-case attorney readings were elicited from the input materials (pre-amendment claim, rejection rationale, cited prior art, and the model-generated amendment) before the automated scores were revealed; the comparison was then made between the attorney's predicted direction and the automated direction. The full $\kappa$ study will pre-register the same blinding protocol on the larger 30--50-case sample, with double-rated readings to estimate inter-rater reliability alongside agreement with the automated channels.

\section{Batch replication protocol}
\label{app:replication}

\textbf{Sampling procedure.} Deterministic cohort selection (\texttt{scripts/select\_cohort.py}) under the six-axis marginal-match procedure described in \S\ref{sec:cohort} and Appendix~\ref{app:cohort}. The procedure is reproducible from seed.

\textbf{Batch 0.} The seed used for primary analysis is 42; cohort case identifiers are released as \texttt{cohort\_batch0.json}.

\textbf{Additional batches.} All results in this paper are reported on batch~0. Because the selection procedure is deterministic given a seed, any additional 100-case cohort can be regenerated with \texttt{select\_cohort.py} under a new seed. We run a single well-constructed cohort by design: pre-emptive multi-batch analysis would either double the budget without informing the primary hypotheses or, if abbreviated, weaken probe depth.

\subsection{Stage 2 escalation criteria}
\label{app:stage2}

Stage 2 (Claude Haiku 4.5 and GPT-4o-mini on the full probe battery) is performed when at least two of the following hold consistently across both Stage-1 flagships: (i) Probe F yields $\Delta\mathrm{C5} > 0.2$ relative to no-retrieval baseline; (ii) Probes F and G yield comparable C5 shifts ($|\Delta_F - \Delta_G| < 0.1$, same sign); (iii) retrieval does not reduce C5 below baseline under at least one of F or G. When fewer than two conditions hold, Stage 1 already answers the question asymmetrically across vendor families, and Stage 2 primarily clarifies scope rather than shifting conclusions.

\section{Stratified results}
\label{app:stratified}

Per-stratum result tables (by statute section, technology center, XML format, and amendment pattern class) are released as supplementary TSVs alongside the corpus and parsing code. The cohort's six-axis marginal match (Appendix~\ref{app:cohort}) ensures each stratum has at least five cases, supporting within-stratum effect-size estimation.

\section{Corpus release index}
\label{app:release}

The released corpus consists of: the parsed JSONL of 7{,}385 four-tuples; the parsing code; the application-number index permitting zero-cost reconstruction from USPTO ODP; and \texttt{cohort\_batch0.json} (the batch~0 case identifiers); additional batches are regenerable from \texttt{select\_cohort.py} under any seed. All released under a permissive open license; USPTO ODP data has no copyright restriction.

\section{Cohort construction}
\label{app:cohort}

\subsection{Why 100 cases}
The cohort size is the smallest $n$ that simultaneously satisfies four constraints: (i) within-model paired comparisons with $>$80\% power at Cohen's $d = 0.3$, $\alpha = 0.05$; (ii) six-axis marginal-match feasibility with minimum cell size 5 (requires $n \geq \sim$90); (iii) probe-depth budget at 7 probes $\times$ 3 reps $\times$ 4 models $= 84$ calls per case (8{,}400 calls before ablations) which is feasible at $n = 100$ but triples at $n = 300$ without adding decisive within-model power; (iv) attorney-validation headroom, since a 30--50-case audit is a meaningful fraction of $n = 100$ but only 3\% of $n = 1{,}000$.

\subsection{Why stratified, not random}
A random 100-case draw from the 4{,}270-case eligibility pool would under-represent the rejection outcomes most informative for H2 contrasts (Reversed cells at $\sim$30\% raw frequency), rare statutes (\S101-related and \S112-alone at $\sim$6--10\% each, falling below the cell-minimum 5 threshold with non-trivial probability), and minority technology centers (long tail of 1--3-case bins that cannot support per-TC stratified robustness). Stratified sampling with target floors guarantees presence on all six axes.

\subsection{Sampling procedure}
Iterative marginal matching, not cross-product enumeration. The six-axis cross-product has too many empty cells at $n = 100$; iterative matching instead scores each case in the pool by the rarity of its six-axis vector relative to the targets, greedily accepts cases that move the cohort toward all six target marginals simultaneously, and uses the seed to break ties. At each step the procedure checks cell-minimum $m = 5$ for every bin of every axis. Code: \texttt{scripts/select\_cohort.py}. Seed for batch 0: 42.

\subsection{Eligibility filter}
The 4{,}270-case pool is the subset of the 7{,}385 $\beta$-parsed corpus satisfying: (i) parse health (pre- and post-claims parseable, at least one CTNF rejection instance, statute section identifiable on the first rejection); (ii) axis coverage (subdecision outcome $\in$ \{Affirmed, Reversed, Affirmed-in-Part\}, excluding dismissed and remanded outcomes that appear in $\sim$8\% of the source pool); (iii) no obvious data defect (pre-claim 1 $\geq$ 50 characters, at least one modified or cancelled per-claim action so that there is something to score).

\subsection{Batch 0: achieved vs. target}
Marginal match per axis: outcome 40/40 Affirmed, 40/40 Reversed, 20/20 AIP. Statute 35/20/15/15/10/5 target vs.\ 36/20/15/15/9/5 achieved across 103-alone, 102+103, 103+112, 101-related, 112-alone, and other. Technology center 20/17/14/12/12 across the top six TCs (exact match). Pattern modify\_only 45 / cancel\_heavy 20 / mixed 15 / new\_or\_add 10 / other 10 (exact match). Format dtd\_legacy 25 / ns\_claims 75 (exact match). Year 25/45/30 target vs.\ 24/46/30 achieved across 2010--14, 2015--19, 2020--25. All bins pass the $m = 5$ minimum (smallest bin: statute "other" at 5).

\subsection{Post-registration amendments}

\textbf{Cohort.} PILOT-2019006614 was dropped because its $\beta$ pre\_clm XML contains two \texttt{<Claim>} wrappers with eleven unnumbered \texttt{<ClaimText>} elements and no inline-header pattern, making pre-amendment text not deterministically recoverable. A pre-declared replacement rule selected the lowest app\_num case in the eligibility pool matching the dropped case's six-axis vector (Affirmed / \S101-related / TC 3600 / 2020--25 / new\_or\_add / ns\_claims) with parse-success under the patched parser. Selected: PILOT-2022000316. Marginal impact: zero.

\textbf{Model matrix.} GPT-4o-mini replaces GPT-5-mini in the smaller-OpenAI cell. Trigger: in the $N{=}10$ multi-model smoke run, GPT-5-mini produced 25 of 30 zero-parseable-claim responses, 13 of 30 max\_tokens truncations, and did not accept the pre-registered temperature 0.3. A pre-declared fallback rule substitutes to the next stable smaller OpenAI model supporting temperature control and canonical claim format. Verification on GPT-4o-mini: 29 of 30 parseable responses, 0 of 30 truncations, temperature 0.3 accepted.

\subsection{Retrieval-pool disjointness}
\label{app:retrieval-pool}

The invariant cohort $\cap$ pool $= \emptyset$ holds by construction, since the pool is the $\beta$ corpus with the 100 cohort cases removed.

\section{Significance tests and equivalence}
\label{app:significance}

For each model we aggregate the three replicates to a per-case mean, pair the retrieval condition against the no-retrieval baseline case-by-case, and apply a paired Wilcoxon signed-rank test. $p$-values are Holm--Bonferroni corrected within the C5 test family; bootstrap 95\% CIs on the median paired difference use 10{,}000 resamples (seed 42). Equivalence against the pre-registered $\pm 0.2$ margin is assessed by two one-sided tests (TOST) on the median $\Delta$C5.

\begin{table}[t]
\centering
\small
\setlength{\tabcolsep}{4pt}
\caption{C5 baseline-versus-retrieval tests. No comparison is significant uncorrected (0 of 8), and none survives Holm correction. Median-based TOST establishes equivalence to zero within the $\pm 0.2$ margin for every model under both probes.}
\label{tab:significance}
\begin{tabular*}{\columnwidth}{@{\extracolsep{\fill}}lcccc@{}}
\toprule
Model & med $\Delta$F & $p_F$ & med $\Delta$G & $p_G$ \\
\midrule
Claude Sonnet~4  & $+0.00$ & 0.52 & $+0.00$ & 0.78 \\
Claude Haiku~4.5 & $+0.07$ & 0.11 & $+0.01$ & 0.33 \\
GPT-5.4          & $+0.00$ & 0.14 & $+0.01$ & 0.30 \\
GPT-4o-mini      & $+0.00$ & 0.87 & $+0.00$ & 0.78 \\
\bottomrule
\end{tabular*}
\end{table}

On C2 the same battery finds 3 of 8 comparisons nominally significant (Haiku~4.5 F $p{=}0.008$ and G $p{=}0.036$; Sonnet~4 F $p{=}0.049$), all decreases, none surviving Holm correction. The effect is carried by Haiku~4.5 rather than a Claude-versus-GPT boundary (\S\ref{sec:family-divergence}).

\section{Paraphrase-sensitive (semantic) C1}
\label{app:semantic-c1}

To check that the grounding null is not an artifact of lexical (trigram) overlap, we recompute C1 with a paraphrase-sensitive embedding scorer (\texttt{all-MiniLM-L6-v2}), scoring the same generated and rejected limitations by embedding cosine rather than token overlap. Lexical and semantic C1 are only loosely concordant (Spearman $\rho = 0.37$) and flag different cases, so the semantic variant is a genuine second measurement rather than a re-derivation. The retrieval null is unchanged: 0 of 8 baseline-versus-retrieval comparisons survive Holm correction under semantic C1, and the two uncorrected near-misses (Haiku~4.5 and GPT-4o-mini under F) are \emph{decreases}, not the grounding gains the optimistic account predicts.

\section{Dense-retriever realization of Probe G}
\label{app:dense-g}
Dense-G is a controlled swap of the retrieval mechanism only: an embedding-cosine top-3 retriever (\texttt{all-MiniLM-L6-v2} over rejection-plus-attacked-claim text) replaces Probe G's deterministic feature matcher, holding the pool, $k{=}3$, exemplar fields, injection format, and base prompt byte-identical to F and G. A shared-space retrieval-quality log (Table~\ref{tab:dense-quality}) confirms the three conditions are genuinely different retrievers.

\begin{table}[t]
\centering \small
\caption{Retrieval quality, F vs G vs dense-G, over each condition's actual picks (300 per condition).}
\label{tab:dense-quality}
\begin{tabular*}{\columnwidth}{@{\extracolsep{\fill}}lccc@{}}
\toprule
Condition & mean cosine & struct.\ (0--3) & statute match \\
\midrule
random-F      & 0.587 & 0.746 & 31.7\% \\
structural-G  & 0.607 & 2.342 & 90.0\% \\
dense-G       & 0.871 & 1.173 & 55.3\% \\
\bottomrule
\end{tabular*}
\end{table}

Dense-G retrieves far more semantically similar exemplars than F or G (cosine 0.87 vs $\sim$0.60) while its structural-tag overlap sits between them, so it is a real third mechanism rather than a re-derivation of G. Generation ran on all four models (1{,}200 calls, 0 failures, \$18.32). Scored with the same deterministic pipeline against a Holm-corrected 16-test family (four models $\times$ C5, C2, lexical C1, semantic C1), 0 of 16 comparisons survive correction and no model shows a significant C5 increase; the closest is Haiku~4.5 on C5 ($p{=}0.053$, uncorrected). The null is thus robust to how the retriever is built, not only to which fixed matcher is chosen.

\section{Retrieval-depth sensitivity}
\label{app:ksweep}
$k{=}3$ was a fixed, untuned default balancing exemplar diversity against prompt length and inference cost. Varying only $k\in\{1,3,5,10\}$ (cohort, retriever, scaffold, exemplar ordering, model, and decoding held fixed), C5 medians (random-F $|$ structural-G) are given in Table~\ref{tab:ksweep}.

\begin{table}[t]
\centering \small \setlength{\tabcolsep}{4pt}
\caption{C5 median by retrieval depth, random-F $|$ structural-G.}
\label{tab:ksweep}
\begin{tabular*}{\columnwidth}{@{\extracolsep{\fill}}lcccc@{}}
\toprule
Model & $k{=}1$ & $k{=}3$ & $k{=}5$ & $k{=}10$ \\
\midrule
Sonnet~4    & 4.90/4.94 & 4.94/4.94 & 4.94/4.92 & 5.00/4.94 \\
Haiku~4.5   & 4.15/4.29 & 4.29/4.27 & 4.11/4.37 & 4.32/4.41 \\
GPT-5.4     & 4.93/4.85 & 4.94/4.83 & 4.93/4.88 & 4.85/4.92 \\
GPT-4o-mini & 4.12/3.95 & 4.27/4.06 & 4.05/4.10 & 4.04/3.94 \\
\bottomrule
\end{tabular*}
\end{table}

C5 shows no monotonic increase with $k$. Quantifying depth variation as the maximum deviation from the $k{=}3$ value, this is $\leq 0.18$ for seven of the eight model$\times$condition series; the one larger case (GPT-4o-mini, random-F, 0.23) is non-monotonic, peaking at $k{=}3$ and lower at $k{=}5$ and $k{=}10$. C1 is practically stable across depths (only Sonnet~4 dips slightly at $k{=}10$); C3 rises mildly with $k$ for the Claude models and is flat for GPT; C2 is noisy with no monotonic trend. Applying the pre-registered criteria at every depth yields the same qualitative H1--H3 verdicts, so the conclusions are not artifacts of $k{=}3$. Prompt length grows with $k$ as expected (approximate median 4.5k tokens at $k{=}1$ to 6.2k at $k{=}5$, \texttt{cl100k} tokenizer), and the office-action truncation rate (8{,}000-character cap on the rejection text) is 80\% and constant across depths.

\section{Truncation and parse robustness}
\label{app:truncation}
The examiner rejection is presented to the model as a set of structured rejection instances (attacked claim numbers, statute section and subsection, rejection type, and cited references; \S\ref{sec:probes}) followed by the verbatim examiner body, and only the body is capped at 8{,}000 characters (Appendix~\ref{app:ksweep}). Across the 100-case cohort the footer-trimmed body exceeds the cap in 80\% of cases (median body length 13{,}448 characters, range 2{,}333--41{,}801), so most prompts truncate the tail of the verbatim reasoning. The truncation does not, however, remove the material the metric scores against: the structured rejection instances precede the body and are never clipped, so the attacked-claim identification, statutory basis, and cited references are present for all 100 cases, and a statutory-rejection statement (``\ldots rejected under 35 U.S.C.\ \ldots'') remains within the retained window for all 100 cases. C1 aligns the generated amendment against the claims named as rejected, which are carried by the structured instances; truncation therefore clips only the tail of the examiner's verbatim prose, not the identification of the attacked limitation.

\paragraph{Differing per-model subsets.} A generated amendment occasionally fails to parse and is dropped, so each model is scored on a slightly different subset. Per-model parse-success is 97.4\% overall (\S\ref{sec:results}). Because failures fall on different cases per model, we recompute the H1 criterion on the common subset of 65 cases parsed by all four models under every condition: no model reaches the $\Delta$C5 $\geq 0.2$ threshold (maximum $+0.10$ for GPT-4o-mini), so the prior-injection verdict is unchanged when every model is scored on identical cases. Dropped cases track the cohort's marginals rather than concentrating in a stratum: by first-rejection statute, \S103 is 40\% of failures vs.\ 36\% of the cohort; by amendment pattern, modify-only is 40\% vs.\ 45\%; and the technology-center distribution is comparable (largest shift: art unit 1700 at 25\% of failures vs.\ 14\% of the cohort).

\paragraph{Complete-rejection subset.} On the 20 cases whose rejection body is not truncated, per-case C5 estimates are dominated by small-sample variance, consistent with the $N$-sensitivity documented in \S\ref{sec:methodology-validity} (where a $N{=}10$ estimate of $+0.84$ reversed to $-0.12$ at $N{=}300$); we therefore do not adjudicate H1 on this 20-case subset. Because truncation leaves the scored attacked-limitation identification intact, there is no mechanism by which it would bias the verdict, and the common-subset check above provides the properly-powered robustness result.

\end{document}